%% file: main.tex
\documentclass{sharpa}
\usepackage{amsmath}
\usepackage{enumerate} 
\usepackage{algorithm    }
\usepackage{algpseudocode}
\usepackage{amsfonts}
\usepackage{amsthm}
\usepackage{newtxtt}
\usepackage{diagbox} 
\usepackage{colortbl}
\usepackage{amssymb}
\usepackage{xspace}
\usepackage{wrapfig}
\usepackage{adjustbox}
\usepackage{tabularx}
\usepackage{booktabs}
\usepackage{mathtools}
\usepackage{tikz}
\usepackage{enumitem}
\usepackage{silence}
\usepackage{dsfont}
\usepackage[table]{xcolor}
\usepackage[dvipsnames]{xcolor}
\usepackage{multirow}
\usepackage{makecell}
\usepackage{xfakebold}
\usepackage{natbib}
\usepackage{siunitx}
\usepackage{cleveref}

\input{math_commands}

\input{macros_arxiv}

\usepackage{needspace}
\newsavebox{\sidecapbox}
\newsavebox{\sidetallbox}

\newcommand{\method}{SALA\xspace}

\makeatletter
\DeclareRobustCommand\onedot{\futurelet\@let@token\@onedot}
\def\@onedot{\ifx\@let@token.\else.\null\fi\xspace}
\makeatother
 
\def\ie{\emph{i.e}\onedot}

\definecolor{oursrow}{rgb}{0.90,0.94,0.98}

\definecolor{tgtblue}{RGB}{39,54,194}
\definecolor{srcorange}{rgb}{0.90,0.45,0.05}
\newcommand{\tgt}[1]{{\color{tgtblue}#1}}
\newcommand{\src}[1]{{\color{srcorange}#1}}

\definecolor{textgray}{HTML}{6E6E73}
\usetikzlibrary{positioning, calc}
\usetikzlibrary{decorations.pathmorphing}

\makeatletter
\patchcmd{\wrong@fontshape}{\@gobbletwo}{}{}{}
\makeatother
\numberwithin{equation}{section} 
\makeatletter
\AtBeginDocument{
  \urlstyle{sf}
  
}
\makeatother

\floatstyle{plain}
\newfloat{algorithmgroup}{tbp}{loag}
\floatname{algorithmgroup}{Algorithm Group}

\renewcommand{\eqref}[1]{\textup{(\ref{#1})}}

\definecolor{light}{RGB}{125, 125, 125}
\crefname{tcb@cnt@pbox}{code}{code}
\Crefname{tcb@cnt@pbox}{Code}{Code}
\crefname{assumption}{assumption}{assumption}
\Crefname{assumption}{Assumption}{Assumptions}

\newtcolorbox[auto counter]{pbox}[2][]{
  colback=white,
  title=Code~\thetcbcounter: #2,
  #1,fonttitle=\sffamily,
  fontupper=\sffamily,
  arc=2pt,
  colframe=bgcolor,
  coltitle=fgcolor,
  colbacktitle=bgcolor,
  toptitle=0.25cm,
  bottomtitle=0.125cm
}

\makeatletter
\newcommand\applefootnote[1]{%
  \begingroup
  \renewcommand\thefootnote{}%
  \renewcommand\@makefntext[1]{\noindent##1}%
  \footnote{#1}%
  \addtocounter{footnote}{-1}%
  \endgroup
}
\makeatother

\definecolor{cverbbg}{gray}{0.90}

\title{SALA: Semantically Aligned Latent Actions Enable Dual-Axis Scaling of Embodied Foundation Models}

\author[1,2,*]{Jiachun Jin}
\author[1,2,*]{Zetong Zhou}
\author[1]{Haonan Wang}
\author[2]{Zhijie Deng}

\affiliation[1]{Sharpa Robotics}
\affiliation[2]{SJTU-Sharpa Joint Lab for Embodied AI, SJTU}

\contribution[*]{\sffamily Equal contribution}

\abstract{
\input{sections/0_abstract}
}

\metadata[Project Website]{\url{https://sala-vla.github.io/}}
\metadata[Correspondence]{\href{mailto:zhijied@sjtu.edu.cn}{\nolinkurl{zhijied@sjtu.edu.cn}}}

\renewcommand{\titlelogowidth}{2.8cm}
\renewcommand{\titlelogoraise}{-16.7pt}

\begin{document}

\maketitle

\input{sections/1_intro}
\FloatBarrier
\input{sections/2_method}
\FloatBarrier
\input{sections/3_exp}
\FloatBarrier
\input{sections/3_related}

\input{sections/4_conclusion}

\FloatBarrier
\setlength{\bibsep}{5pt}
\bibliography{reference}
\bibliographystyle{plainnat}

\newpage
\beginappendix
\input{sections/X_suppl}
\end{document}

%% file: math_commands.tex
\usepackage{amsmath,amsfonts,bm}

\def\eqref#1{equation~\ref{#1}}

\def\1{\bm{1}}

\DeclareMathAlphabet{\mathsfit}{\encodingdefault}{\sfdefault}{m}{sl}
\SetMathAlphabet{\mathsfit}{bold}{\encodingdefault}{\sfdefault}{bx}{n}

%% file: macros_arxiv.tex
\usepackage{parskip}

\usepackage{natbib}
\usepackage{amsfonts,bm}

\usepackage{amsmath, amssymb,mathrsfs, amsthm, mathtools}
\usepackage{xspace}
\usepackage{enumitem}
\usepackage{lipsum}
\usepackage{xpatch}
\usepackage{mathtools}
\usepackage{dsfont}
\usepackage{pgf,tikz}
\usetikzlibrary{positioning,matrix,patterns}
\usepackage{caption}
\usepackage{graphicx}
\usepackage{makecell}
\usepackage{multirow}
\usepackage[mathscr]{euscript}
\definecolor{hanblue}{rgb}{0.27, 0.42, 0.81}
\definecolor{deepred}{HTML}{900C3F}
\definecolor{deepgreen}{HTML}{2F6960}
\hypersetup{
    colorlinks=true,
    linkcolor=hanblue,
    urlcolor=hanblue,
    citecolor=hanblue,
    anchorcolor=hanblue}

\newtheoremstyle{sansserif}{3pt}{3pt}{\normalfont}{}{\sffamily\bfseries}{.}{ }{}

\theoremstyle{sansserif}

\theoremstyle{definition}

\theoremstyle{sansserif}

\theoremstyle{remark}

\DeclarePairedDelimiter\abs{\lvert}{\rvert}%
\DeclarePairedDelimiter\norm{\lVert}{\rVert}%

\makeatletter
\let\oldabs\abs
\def\abs{\@ifstar{\oldabs}{\oldabs*}}
\let\oldnorm\norm
\def\norm{\@ifstar{\oldnorm}{\oldnorm*}}
\makeatother

%% file: sections/0_abstract.tex
Embodied foundation models hinge on diverse robot interaction data, which are costly to collect. Egocentric human videos offer a complementary source of experience, and latent action models (LAMs) provide a popular means of leveraging these videos for policy supervision. Compared with world-action models (WAMs), LAMs can offer greater training efficiency and potentially lower inference costs by avoiding explicit pixel prediction during policy learning and execution. However, most existing LAMs learn latent actions solely through video reconstruction, which can cause their representations to cluster by visual appearance rather than by the underlying actions. As a result, they struggle to align human actions with their robot counterparts.
To bridge the gap, we introduce \method, which emphasizes establishing the semantic correspondence for latent actions. 
Without requiring paired data, we use the language descriptions of unpaired human and robot clips to estimate semantic similarities and supervise cross-embodiment correspondence. Specifically, we translate these similarities into soft neighborhood structures that guide human latent actions toward semantically related robot latents, while robot action reconstruction grounds the latent space in control. 
The resulting shared latent action space provides a unified supervision signal for training an embodied foundation model on both human and robot videos. Notably, latent action supervision enables scaling along two axes: (1) pre-training on increasing amounts of general human video consistently reduces robot action prediction error on downstream tasks, and (2) post-training on increasing amounts of task-specific human video substantially improves real-world robot task success rates. 

%% file: sections/1_intro.tex
\section{Introduction}
\label{sec:intro}

\begin{figure*}[t]
    \centering
    \includegraphics[width=\textwidth]{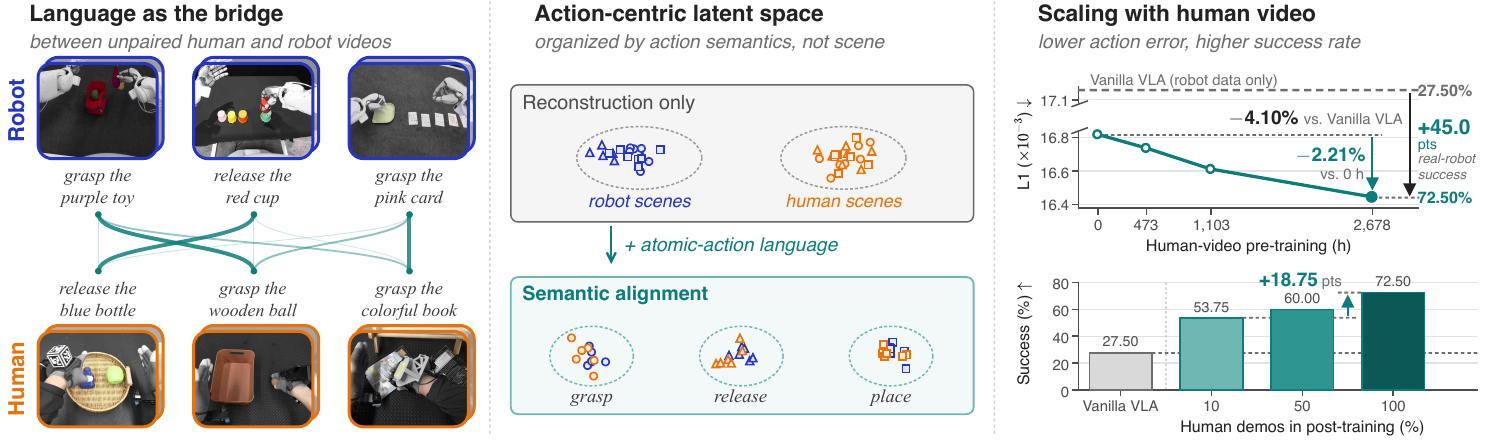}
    \caption{\textbf{Overview of \method.} \textbf{Left:} unpaired human and robot videos are related only through language descriptions of their atomic actions. \textbf{Middle:} supervising a latent action tokenizer with these relations organizes latent actions by action semantics rather than by scene. \textbf{Right:} through the aligned latent actions, human videos supervise the same policy that controls the robot: more human-video pre-training lowers the validation error of robot action prediction without additional robot data, and more task-specific human video in post-training raises the real-world success rate.}
    \label{fig:teaser}
\end{figure*}

Embodied foundation models, such as vision-language-action (VLA) models, connect visual observations and language instructions to robot actions, offering a path toward general-purpose robot control~\citep{black2024pi_0,brohan2023rt,kim2024openvla}.
Their generalization depends on diverse interaction experience across tasks, objects, and environments, yet collecting such data through robot teleoperation remains costly.
Egocentric human videos provide a complementary source of experience, capturing everyday manipulation without requiring robot hardware or teleoperation~\citep{Grauman_2022_CVPR,hoque2026egodex,punamiya2026egoverse}.
Using this experience for policy learning, however, requires bridging the human-to-robot embodiment gap: the same operation can involve different motions and control mechanisms across humans and robots, and is often performed in scenes that differ from the robot's workspace.

Latent action models (LAMs) are a popular way to incorporate human videos into robot policy training without explicit action labels~\citep{ye2025latent,chen2024moto,bu2025univla}.
By encoding visual transitions in a shared latent space, they provide a common prediction target for human and robot experience.
Compared with world-action models (WAMs), which predict future video frames together with actions~\citep{wu2023gr1,yuan2026fastwam,li2026lingbotva}, LAMs avoid explicit pixel prediction during policy learning, which can make training more efficient.
However, most existing LAMs learn latent actions solely through video reconstruction, which constrains each latent to explain the transition within its own clip but does not relate human and robot executions of the same behavior.
Their latents therefore tend to cluster by visual appearance, such as the scene, rather than by the underlying action, so semantically similar human and robot actions need not lie close together.
Aligning human actions with their robot counterparts thus requires supervision that connects their meanings across embodiments.

Our insight is that language can provide this supervision.
Even when human and robot clips are unpaired, natural-language descriptions of their atomic actions, such as ``grasp the cup'', relate a human clip to multiple robot clips with different degrees of semantic similarity, as illustrated on the left of \cref{fig:teaser}.
We propose \method, which trains a latent action tokenizer by translating these similarities into soft neighborhood structures that guide human latent actions toward semantically related robot latents.
Robot latents, grounded in control by action reconstruction, serve as anchors, and the alignment moves only human latents toward them.
As the middle of \cref{fig:teaser} shows, the resulting latent actions are organized by action semantics rather than by scene, giving human videos a prediction target that is meaningful for robot control.

Once trained, the tokenizer labels videos from their frames alone, so its latent actions provide a unified supervision signal for training the policy on both human and robot videos, without action labels or atomic-action descriptions.
This makes human video a scalable source of supervision along two axes.
In pre-training, the policy can absorb large-scale general human videos spanning diverse everyday tasks.
In post-training, the policy can further learn from task-specific human videos, \ie, videos of humans performing the target task.
Because their latent actions are aligned with those of the robot, what the policy learns from these videos transfers directly to robot control.
Since recording such videos requires neither a robot nor teleoperation, they offer a cheap way to improve the policy on a new task beyond what a fixed budget of teleoperated demonstrations provides.

We evaluate \method in simulation and on a real robot.
On RoboTwin 2.0~\citep{chen2025robotwin}, where videos of another robot arm stand in for human videos, \method achieves a higher average success rate than existing LAMs and WAMs trained on the same data.
Analyses of the tokenizer show that language alignment organizes latent actions by atomic action semantics rather than by scene.
On a real robot controlled in a 63-DoF action space, including two 22-DoF dexterous hands, human video helps along both axes.
Pre-training on more general human video, up to 2{,}678 hours, consistently reduces the robot's action prediction error on downstream tasks without additional robot data.
With teleoperated demonstrations fixed, post-training on more task-specific human video raises the average real-robot success rate from 53.75\% to 72.5\%.
The right of \cref{fig:teaser} shows both trends.

Our contributions are as follows:
\begin{itemize}
    \item We use language as relational supervision: atomic-action descriptions align human and robot latent actions from unpaired videos, with the alignment anchored on robot latents grounded in control.
    \item Through the aligned latent actions, we supervise the robot policy with both general human videos and human demonstrations of the target task.
    \item Experiments in simulation and on a 63-DoF real robot show that \method outperforms prior LAMs and WAMs trained on the same data, and that both uses of human video scale: more general human video keeps lowering the validation error of robot action prediction, and more task-specific human video raises real-world success rates.
\end{itemize}

%% file: sections/2_method.tex
\section{Method}
\label{sec:method}
\begin{figure*}[t]
    \centering
    \includegraphics[width=\textwidth]{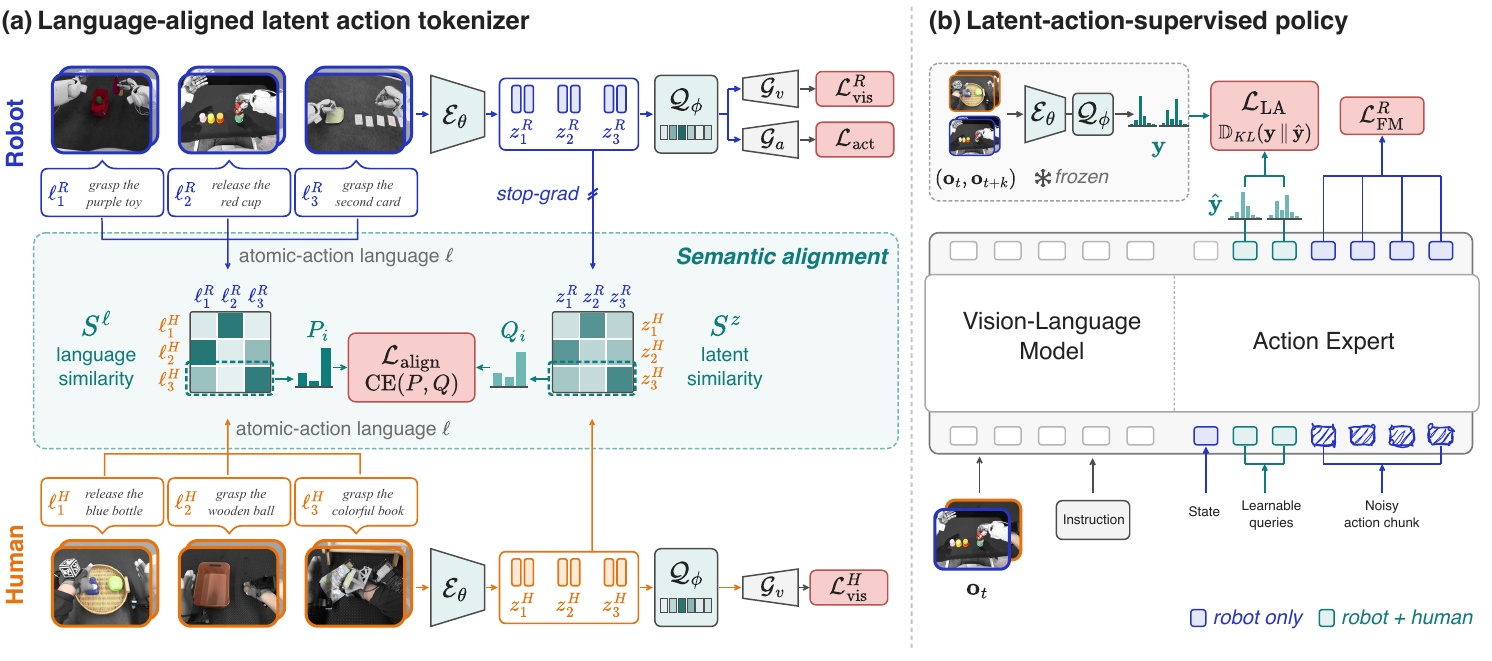}
    \caption{\textbf{Method overview.}
    \textbf{(a)~Latent action tokenizer.} Robot (blue) and human (orange) clips share the visual encoder $\mathcal{E}_\theta$, the SoftVQ quantizer $\mathcal{Q}_\phi$, and the visual decoder $\mathcal{G}_v$; only robot clips train the action decoder $\mathcal{G}_a$. Language similarities $S^{\ell}$ between atomic-action descriptions define soft neighborhoods $P$, which supervise the latent neighborhoods $Q$ derived from latent similarities $S^{z}$, with gradients stopped through robot latents.
    \textbf{(b)~VLA policy.} The frozen tokenizer labels each clip with latent action tokens $\mathbf y$, each a categorical distribution over the codebook. Latent queries in the action expert predict $\hat{\mathbf y}$ through a latent head, supervised by $\mathcal L_{\mathrm{LA}}$ for both human and robot data; robot data additionally train flow matching with $\mathcal L_{\mathrm{FM}}^{R}$. In the action expert, blue inputs (state and noisy actions) are used only for robot data, whereas green latent queries are used for both human and robot data.}
    \label{fig:method}
\end{figure*}

Our goal is to use action-free egocentric human videos, which can be collected at scale, to facilitate the training of robot policies.
We connect these videos to robot control through latent actions.
This connection requires latent actions to establish semantic correspondence across embodiments, which a shared latent space trained only by reconstructing each clip does not guarantee.
As outlined in \cref{fig:method}, \method therefore learns a latent action tokenizer aligned across embodiments by language supervision from atomic-action descriptions, and uses it to leverage human videos in robot policy training.

\subsection{Preliminaries: Latent Action Models}
\label{sec:prelim}
LAMs learn compact representations of visual transitions without action labels by coupling inverse and forward dynamics~\citep{NEURIPS2025_d7b3cef7}.
Given observations $\mathbf o_t$ and $\mathbf o_{t+k}$ separated by a horizon $k$, an inverse-dynamics encoder infers a latent action ${\mathbf z=\mathcal{E}_\theta(\mathbf o_t,\mathbf o_{t+k})}$ that encodes the transition between the two observations.
An information bottleneck produces ${\widetilde{\mathbf z}=\mathcal{Q}_\phi(\mathbf z)}$, limiting how much information from the future observation reaches the decoder.
The forward-dynamics decoder $\mathcal{G}_v$ then reconstructs $\mathbf o_{t+k}$ from the current observation $\mathbf o_t$ and the compressed code $\widetilde{\mathbf z}$.
Training minimizes the reconstruction error in image or feature space over observed pairs~\citep{ye2025latent, oquab2024dinov2, kingma2013auto}.

The inferred codes can serve as prediction targets for policy learning from action-free videos~\citep{ye2025latent,bu2025univla}.
Their usefulness, however, depends on what they encode: accurate reconstruction does not guarantee that the latent captures controllable actions, since it may also retain scene information or changes unrelated to control~\citep{NEURIPS2025_d7b3cef7}.
For human-to-robot transfer, the representation must additionally connect semantically related behaviors across embodiments, a relation that reconstruction within individual clips does not explicitly constrain.
We therefore augment this predictive formulation with robot action supervision and language-guided cross-embodiment alignment.

\subsection{Grounding Latent Actions in Robot Control}
\label{sec:latent_representation}
We train the latent action tokenizer on a human dataset $\src{\mathcal{D}_{H}}$ and an independently collected robot dataset $\tgt{\mathcal{D}_{R}}$:
\begin{equation}
    \src{\mathcal{D}_{H}} = \{(V_i^{H},\ell_i^{H})\}_{i=1}^{N_H}, \quad \tgt{\mathcal{D}_{R}} = \{(V_j^{R},\ell_j^{R},\mathbf{s}_j,\mathbf{a}_j)\}_{j=1}^{N_R}.
\end{equation}
Each clip $V=(\mathbf o_t,\mathbf o_{t+k})$ is represented by its endpoint observations, and $\mathbf{s}$ and $\mathbf{a}$ are the initial robot state and the subsequent action chunk, available only for robot data.
The natural-language description $\ell$ specifies the atomic actions performed by the embodiment in the clip; \cref{sec:supp_robotwin_annotation,sec:supp_real_annotation} describe how these descriptions are obtained in our experiments.
As shown in \cref{fig:method}(a), human and robot clips share the visual encoder $\mathcal{E}_\theta$, the quantizer $\mathcal{Q}_\phi$, and the visual decoder $\mathcal{G}_v$, trained with visual reconstruction losses as in \cref{sec:prelim}.

Visual reconstruction alone does not ensure that the latent captures controllable actions.
Teleoperation data, however, already provide ground-truth robot actions, which we exploit during tokenizer training.
A robot-only action decoder $\mathcal{G}_a$ reconstructs the action chunk $\mathbf a$ from $\widetilde{\mathbf z}$ and the initial state $\mathbf s$, with loss $\mathcal L_{\mathrm{act}}$, which grounds robot latents directly in control~\citep{chen2026iclr-villax}.
Human clips share this space without action labels; the control-grounded robot latents later serve as anchors for cross-embodiment alignment.

\subsection{Robot-Anchored Semantic Transfer}
\label{sec:semantic_transfer}
We align human latent actions toward the robot representations grounded by action reconstruction in \cref{sec:latent_representation}.
These robot representations serve as control anchors, while atomic-action descriptions determine how each human clip should relate to them.

Atomic-action descriptions characterize what a clip does independently of the embodiment performing it.
We therefore encode $\ell$ with a frozen text encoder and use the similarity between language descriptions to measure how semantically related a human clip is to each robot clip.
Because human and robot clips are collected independently, a human clip rarely has an exact robot counterpart.
It can nonetheless relate to several robot clips to different degrees; for example, grasping a cup is closer to grasping a bottle than to placing a cup.
We therefore describe each human clip by a soft neighborhood over the robot clips in a batch, given by a softmax over its language similarities, and use it to supervise the clip's neighborhood in latent space.
\Cref{fig:method}(a) illustrates this with $S^{\ell}$ and $S^{z}$, the language and latent similarities between the human and robot clips in a batch: the dashed rows give the two neighborhoods of one human clip.
Formally, the alignment loss is
\begin{equation}
    \mathcal{L}_{\mathrm{align}}
    = \mathbb{E}_i\Big[\operatorname{CE}\big(
    \underbrace{\operatorname{softmax}_j(s^{\ell}_{ij})}_{P_i},\,
    \underbrace{\operatorname{softmax}_j(s^{z}_{ij})}_{Q_i}
    \big)\Big],
    \label{eq:semantic_alignment}
\end{equation}
where $s^{\ell}_{ij}$ and $s^{z}_{ij}$ are the entries of $S^{\ell}$ and $S^{z}$ for human clip $i$ and robot clip $j$, with $s^{z}_{ij}=\operatorname{sim}(\mathbf z_i^{H},\operatorname{sg}(\mathbf z_j^{R}))$; the softmax temperatures are omitted for clarity.
Here $\operatorname{sim}$ denotes cosine similarity averaged over tokens, and the stop-gradient $\operatorname{sg}$ makes alignment move human latents toward the control-grounded robot anchors rather than pulling the anchors toward human latents, which lack action supervision.
Language thus supervises what visual and action reconstruction leave unconstrained: the semantic correspondence between human and robot latent actions.

The choice of quantizer matters for this alignment.
The relations in $P_i$ are graded, so alignment needs to move the pre-quantization latents $\mathbf z$ continuously, while the policy's targets are the tokens that $\mathcal{Q}_\phi$ produces from them.
A standard hard VQ~\citep{vandenoord2017neurips-neural} fits these graded relations poorly: its commitment loss pulls each $\mathbf z$ toward the selected codeword, and its token, a one-hot assignment to a single codeword, changes only when $\mathbf z$ switches to another code.
We therefore instantiate $\mathcal{Q}_\phi$ with SoftVQ~\citep{Chen_2025_CVPR}, which represents each token as a categorical distribution over the codebook rather than a one-hot assignment, and forms $\widetilde{\mathbf z}$ from the codewords weighted by these distributions; we denote the tokens of a clip by $\mathbf y$.
SoftVQ requires no commitment loss, and its distributions vary smoothly with $\mathbf z$, so the tokens can express graded differences between latents.
As shown in \cref{sec:ablation}, the alignment loss decreases with SoftVQ but stays near its initial value with hard VQ.

The tokenizer is trained with the three signals described above: the visual reconstruction losses $\mathcal{L}_{\mathrm{vis}}^{H}$ and $\mathcal{L}_{\mathrm{vis}}^{R}$ on human and robot clips capture visual transitions, the action reconstruction loss $\mathcal{L}_{\mathrm{act}}$ grounds robot latents in control, and the alignment loss $\mathcal{L}_{\mathrm{align}}$ supervises the correspondence between human and robot latents.
With loss weights $\lambda_{\mathrm{act}}$ and $\lambda_{\mathrm{align}}$, the overall objective is
\begin{equation}
    \mathcal{L}_{\mathrm{tok}}
    = \frac{1}{2}\left(
    \mathcal{L}_{\mathrm{vis}}^{H}
    + \mathcal{L}_{\mathrm{vis}}^{R}\right)
    + \lambda_{\mathrm{act}}\mathcal{L}_{\mathrm{act}}
    + \lambda_{\mathrm{align}}\mathcal{L}_{\mathrm{align}}.
    \label{eq:tokenizer_objective}
\end{equation}

\subsection{Latent-Action-Supervised Policy Learning}
\label{sec:policy_learning}
After training, the tokenizer is frozen and serves as a teacher that labels human and robot clips with latent actions.
For a clip $(\mathbf o_t,\mathbf o_{t+k})$, the teacher outputs its latent action tokens $\mathbf y$, each a categorical distribution over the codebook.
We keep these distributions rather than discretizing them into code indices, retaining the graded similarity structure of the latent space.
The future observation $\mathbf o_{t+k}$ is used only to compute $\mathbf y$; the policy itself observes only the current observation, so neither the teacher nor future frames are needed at deployment.

As shown in \cref{fig:method}(b), our policy adopts a $\pi_0$-style architecture~\citep{black2024pi_0}: a vision-language model (VLM) encodes observations and language, and an action expert generates action chunks by flow matching.
We add learnable latent-action queries to the action expert, which predict the token distributions $\hat{\mathbf y}$ through a latent head, and let the noisy action tokens attend to these queries, so that latent-action prediction directly conditions action generation.
For human videos, which provide neither states nor actions, the action expert processes only the latent queries.

Both human and robot data are supervised by the latent-action loss $\mathcal L_{\mathrm{LA}}=\mathbb{D}_{\mathrm{KL}}(\mathbf y\,\|\,\hat{\mathbf y})$, the KL divergence between the teacher's and the predicted categorical distributions over the codebook, averaged over tokens; robot data additionally contribute the flow-matching action loss $\mathcal L^R_{\mathrm{FM}}$:
\begin{equation}
    \mathcal L_{\mathrm{policy}}
    = \mathcal L_{\mathrm{FM}}^{R}
    + \lambda_{\mathrm{LA}}\big(\mathcal L_{\mathrm{LA}}^{R} + \mathcal L_{\mathrm{LA}}^{H}\big).
    \label{eq:policy_objective}
\end{equation}
Human videos thus train the same VLM and action expert that generate robot actions.

We train the policy in stages that follow the availability of data.
In pre-training, the policy learns from the action-free videos in $\src{\mathcal D_H}$ with $\mathcal L_{\mathrm{LA}}^{H}$ alone, predicting the latent actions that the language-aligned teacher assigns to them.
Since the teacher requires only video frames, this stage needs neither action labels nor atomic-action descriptions, and can therefore scale with the amount of available human video.
In mid-training, it is trained on target-robot teleoperation data $\tgt{\mathcal D_R}$ with $\mathcal L_{\mathrm{FM}}^{R}+\lambda_{\mathrm{LA}}\mathcal L_{\mathrm{LA}}^{R}$, grounding these latent-action predictions in robot actions.
When task-specific data are available, an optional post-training stage uses both task-specific teleoperated and human demonstrations with the full objective in \cref{eq:policy_objective}.
Human demonstrations of the target task are cheap to collect, since they require no robot, and the aligned latent space lets them supervise the same action expert that controls the robot.
Each experiment in \cref{sec:exp} specifies which stages it uses and on which data; architectural details and loss weights are given in \cref{sec:supp_impl}.

%% file: sections/3_exp.tex
\begin{figure*}[!tb]
    \centering
    \includegraphics[width=\textwidth]{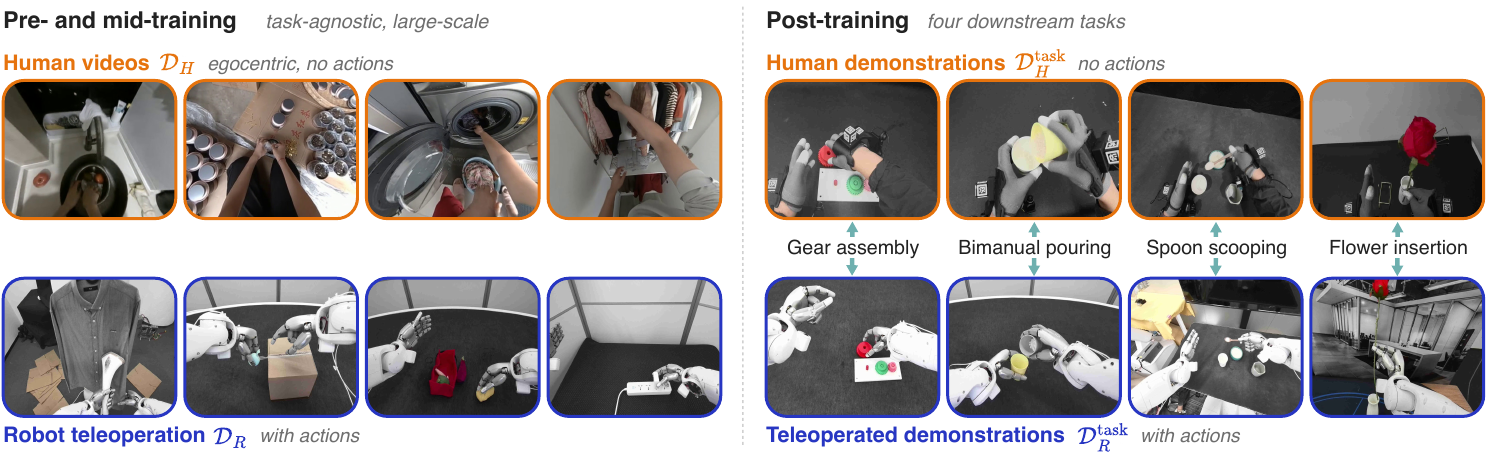}
    \caption{\textbf{Real-world data.} Left: examples of the large-scale data used in pre- and mid-training, \ie, egocentric human videos $\src{\mathcal D_H}$ and robot teleoperation data $\tgt{\mathcal D_R}$. Right: the four downstream tasks, each performed by a person, recorded as human demonstrations $\src{\mathcal D_H^{\mathrm{task}}}$, and by the robot under teleoperation, recorded as teleoperated demonstrations $\tgt{\mathcal D_R^{\mathrm{task}}}$.}
    \label{fig:real_data}
\end{figure*}

\section{Experiments}
\label{sec:exp}
Our experiments test the central claim of \method: language-aligned latent actions turn videos of other embodiments into supervision for robot policies that is both \emph{effective} and \emph{scalable}.
After summarizing the implementation in \cref{sec:impl}, we evaluate the two properties in turn.
For effectiveness, \cref{sec:sim} compares \method on a standardized simulation benchmark with existing LAMs and WAMs trained on the same cross-embodiment data.
For scalability, \cref{sec:real} asks whether \method keeps improving the target robot in the real world as more human video is used, both with increasing human-video pre-training budgets and in task-specific post-training.
We then ablate the design choices of our tokenizer and analyze whether its latent actions are aligned across embodiments in \cref{sec:ablation}.

\subsection{Implementation Details}
\label{sec:impl}
We summarize the main configuration here.
The tokenizer encodes each clip, whose two observations are separated by a dataset-specific horizon $k$, into 8 latent tokens, quantized by SoftVQ with a codebook of 256 codes; visual reconstruction targets frozen DINOv2~\citep{oquab2024dinov2} features of the future observation, and atomic-action descriptions are embedded by a frozen text encoder, CLIP ViT-B/32 in simulation and Qwen3-Embedding-4B~\citep{qwen3embedding} in the real world.
The policy uses Qwen3-VL-4B~\citep{Qwen3-VL} as the VLM and a separately parameterized Qwen3-style transformer with 0.6B parameters as the action expert, which attends jointly with the VLM at every layer.
The VLM is initialized from the pretrained Qwen3-VL-4B-Instruct weights and the action expert from scratch, and each subsequent training stage starts from the checkpoint of the previous one.
Experiment-specific tokenizer settings, loss weights, and further architectural details are given in \cref{sec:supp_impl}.

\subsection{Effectiveness in Simulation}
\label{sec:sim}
We first compare \method with existing approaches to cross-embodiment video pre-training on a standardized simulation benchmark.
Specifically, we ask whether, given the same videos of a non-target embodiment, \method achieves higher target-robot success rates than other LAMs and WAMs.

\begin{table}[!tb]
    \centering
    \sbox{\sidecapbox}{\begin{minipage}[t]{0.525\textwidth}
    \centering
    \caption{\textbf{Training data.} Following \cref{sec:latent_representation}, $\src{\mathcal D_H}$ contains only videos of a \src{non-target embodiment}, whereas $\tgt{\mathcal D_R}$ additionally provides the states and actions of the \tgt{target robot}. Lang.: annotated with atomic-action descriptions, which are used only for tokenizer training; in the real world, only the subset $\src{\mathcal D_H^{\mathrm{tok}}}\subset\src{\mathcal D_H}$ is annotated. The per-task demonstrations $\tgt{\mathcal D_R^{\mathrm{task}}}$ and $\src{\mathcal D_H^{\mathrm{task}}}$ used in real-world post-training are listed in \cref{tab:tasks}. Human-video pre-training hours denote equivalent training exposure.}
    \label{tab:data}
    \end{minipage}}%
    \sbox{\sidetallbox}{\begin{minipage}[t]{0.455\textwidth}
        \centering
        \caption{\textbf{Success rates (\%) on RoboTwin 2.0.} WAMs and LAMs are additionally trained on videos from $\src{\mathcal D_H}$ and $\tgt{\mathcal D_R}$. PT.: initialized from the official checkpoint pre-trained on large-scale robot data.}
        \label{tab:robotwin}
        \small
        \setlength{\tabcolsep}{3pt}
        \begin{tabular}{lcccc}
            \toprule
            Method & PT. & Clean & Random & Avg.\ \\
            \midrule
            \multicolumn{5}{l}{\textit{VLAs}} \\
            $\pi_{0.5}$ & \checkmark & 70.7 & 46.6 & 58.7 \\
            X-VLA & \checkmark & 67.9 & 20.9 & 44.4 \\
            \midrule
            \multicolumn{5}{l}{\textit{WAMs}} \\
            FastWAM & $\times$ & 79.4 & 3.2 & 41.3 \\
            LingBot-VA & \checkmark & 75.8 & 38.4 & 57.1 \\
            InternVLA-A1.5 & $\times$ & 67.8 & 40.1 & 54.0 \\
            OpenWAM & $\times$ & 86.8 & 3.9 & 45.4 \\ 
            \midrule
            \multicolumn{5}{l}{\textit{LAMs}} \\
            UniVLA & \checkmark & 48.6 & 5.6 & 27.1 \\
            UniT & $\times$ & 47.5 & 7.2 & 27.4 \\
            RotVLA & \checkmark & 74.4 & 28.0 & 51.2 \\
            \midrule
            \rowcolor{oursrow} \method & $\times$ & 80.2 & 48.6 & 64.4 \\
            \bottomrule
        \end{tabular}
        \end{minipage}}%
    \begin{minipage}[t][\dimexpr\ht\sidetallbox+\dp\sidetallbox\relax][s]{0.525\textwidth}
    \centering
    \usebox{\sidecapbox}\par
    \vfill
    \small
    \setlength{\tabcolsep}{3pt}
    \renewcommand{\arraystretch}{1.2}
    \begin{tabular}{llccrr}
        \toprule
        Data & Embodiment & Actions & Lang. & Trajectories & Hours \\
        \midrule
        \multicolumn{6}{l}{\textit{Simulation (\cref{sec:sim})}} \\
        $\src{\mathcal D_H}$ & \src{ARX-X5} & $\times$ & \checkmark & 27{,}000 & 68.0 \\
        $\tgt{\mathcal D_R}$ & \tgt{Aloha-AgileX} & \checkmark & \checkmark & 2{,}500 & 9.2 \\
        \midrule
        \multicolumn{6}{l}{\textit{Real world (\cref{sec:real})}} \\
        $\src{\mathcal D_H}$ & \src{Human} & $\times$ & partial & 71{,}582 & 2{,}678 \\
        $\src{\mathcal D_H^{\mathrm{tok}}}$ & \src{Human} & $\times$ & \checkmark & 27{,}406 & 983 \\
        $\tgt{\mathcal D_R}$ & \tgt{Sharpa North} & \checkmark & \checkmark & 70{,}859 & 888 \\
        \bottomrule
    \end{tabular}\par\kern0pt
    \end{minipage}\hfill
    \usebox{\sidetallbox}
\end{table}

\paragraph{Setup.}
We conduct this comparison on RoboTwin 2.0~\citep{chen2025robotwin}, a bimanual manipulation benchmark that provides expert trajectories for 50 dual-arm tasks across five robot embodiments.
We take \tgt{Aloha-AgileX} as the target embodiment and \src{ARX-X5} as the non-target embodiment.
As listed in \cref{tab:data} and illustrated in \cref{fig:robotwin_embodiments}, $\src{\mathcal D_H}$ keeps only the videos of 27{,}000 \src{ARX-X5} trajectories from both clean and random scenes, whereas $\tgt{\mathcal D_R}$ contains only 2{,}500 complete \tgt{Aloha-AgileX} trajectories from clean scenes.
This order-of-magnitude gap mimics the practical regime in which target-robot data are scarce while videos of other embodiments are abundant.
Clips in both datasets are further annotated with atomic-action descriptions, which are used only to train our tokenizer; the annotation procedure and examples are given in \cref{sec:supp_robotwin_annotation}.
We evaluate on \tgt{Aloha-AgileX} in both the clean and domain-randomized settings of the benchmark, denoted Clean and Random.
In each setting, we run 100 rollouts per task and report the success rate averaged over all 50 tasks.

\paragraph{Training.}
The tokenizer is trained on $\src{\mathcal D_H}$ and $\tgt{\mathcal D_R}$ with \cref{eq:tokenizer_objective}, where only \tgt{Aloha-AgileX} clips contribute action reconstruction.
Following \cref{sec:policy_learning}, the policy is then pre-trained on the \src{ARX-X5} videos to predict latent actions only, and mid-trained on $\tgt{\mathcal D_R}$ with both the flow-matching action loss and the latent-action loss.
We omit task-specific post-training in this setting.
The tokenizer settings of this experiment are listed in \cref{tab:supp_tokenizer}.

\paragraph{Baselines.}
We compare with three types of baselines:
(i) VLAs, \ie, $\pi_{0.5}$~\citep{intelligence2025pi05} and X-VLA~\citep{zheng2025xvla}, trained on $\tgt{\mathcal D_R}$ only;
(ii) WAMs, \ie, FastWAM~\citep{yuan2026fastwam}, LingBot-VA~\citep{li2026lingbotva}, InternVLA-A1.5~\citep{ma2026internvlaa15}, and OpenWAM~\citep{wang2026openwam}, additionally trained with their future-prediction objectives on $\src{\mathcal D_H}$ and $\tgt{\mathcal D_R}$; and
(iii) LAMs, \ie, UniVLA~\citep{bu2025univla}, UniT~\citep{chen2026unit}, and RotVLA~\citep{li2026rotvla}, additionally trained to predict latent actions on $\src{\mathcal D_H}$ and $\tgt{\mathcal D_R}$.

\paragraph{Results.}
\method achieves the highest success rate in Random and the highest average in \cref{tab:robotwin}.
Although it is initialized only from a general-purpose VLM, it outperforms every baseline initialized from a robot-pretrained checkpoint.
None of the other LAMs relates the two embodiments through action semantics, so an \src{ARX-X5} latent and an \tgt{Aloha-AgileX} latent need not encode the same action; what their policies learn from $\src{\mathcal D_H}$ may therefore not transfer to \tgt{Aloha-AgileX}.
\method aligns \src{ARX-X5} latents to \tgt{Aloha-AgileX} latents grounded in control, so pre-training on $\src{\mathcal D_H}$ supervises the action expert with targets that are meaningful for \tgt{Aloha-AgileX} control.
Three of the four WAMs instead predict future videos, a target that also requires modeling scene appearance, whereas InternVLA-A1.5 predicts latent foresight tokens rather than pixels.
Consistent with this, FastWAM and OpenWAM match or exceed \method in Clean but collapse in Random, where domain randomization changes scene appearance.
In contrast, \cref{sec:ablation} shows that language-aligned latent actions are organized by action semantics rather than by scene.

\subsection{Scalability in the Real World}
\label{sec:real}
We next deploy \method on a real robot to study how it scales with egocentric human videos.
We examine two levels: scaling human-video pre-training to 2{,}678 hours, and scaling human demonstrations of each downstream task in post-training.

\paragraph{Setup.}
All real-world experiments are conducted on a Sharpa North wheeled humanoid equipped with two 22-DoF Sharpa Wave dexterous hands, whose 63-dimensional joint-space action covers the arms, hands, torso, and neck; \cref{sec:supp_platform} details the platform.
As summarized in \cref{tab:data}, we use action-free egocentric human videos $\src{\mathcal D_H}$ for 2{,}678 equivalent hours of pre-training, alongside 888 hours of robot teleoperation data $\tgt{\mathcal D_R}$.
Each downstream task further pairs a limited set of teleoperated demonstrations $\tgt{\mathcal D_R^{\mathrm{task}}}$ with human demonstrations $\src{\mathcal D_H^{\mathrm{task}}}$, egocentric videos of a person performing the same task; \cref{fig:real_data} shows the large-scale data on the left and these demonstrations on the right.
Teleoperated demonstrations supply the task's robot actions, but they are costly to collect and hard to scale.
Human demonstrations carry no actions, but since they require no robot, they are cheap to collect and easy to scale.
As in simulation, atomic-action descriptions are used only for tokenizer training, so we annotate only the data used to train it: $\tgt{\mathcal D_R}$ and a subset $\src{\mathcal D_H^{\mathrm{tok}}}\subset\src{\mathcal D_H}$, while pre-training on the full $\src{\mathcal D_H}$ requires no annotation.
We annotate these data with a vision-language model, as detailed in \cref{sec:supp_real_annotation}.

\paragraph{Downstream tasks.}
We evaluate four downstream tasks that cover different manipulation capabilities: \emph{gear assembly} and \emph{flower insertion} require fine-grained manipulation, \emph{bimanual pouring} requires coordination between the two hands, and \emph{spoon scooping} requires tool use.
\Cref{tab:tasks} summarizes each task and \cref{sec:supp_real_tasks} gives further details about the four tasks.

\paragraph{Training.}
The tokenizer is trained on $\src{\mathcal D_H^{\mathrm{tok}}}$ and $\tgt{\mathcal D_R}$ with \cref{eq:tokenizer_objective}.
Following the full recipe of \cref{sec:policy_learning}, the policy is pre-trained on $\src{\mathcal D_H}$ to predict latent actions only, and mid-trained on $\tgt{\mathcal D_R}$ with both the flow-matching action loss and the latent-action loss.
It is then post-trained for each task on $\tgt{\mathcal D_R^{\mathrm{task}}}$ and $\src{\mathcal D_H^{\mathrm{task}}}$ together, with the human demonstrations supervising latent-action prediction only.
The tokenizer settings of this experiment are listed in \cref{tab:supp_tokenizer}.

\begin{table*}[!tb]
    \centering
    \sbox{\sidetallbox}{\begin{minipage}[t]{0.412\textwidth}
    \centering
    \caption{\textbf{Downstream tasks.} Teleoperated demonstrations $\tgt{\mathcal D_R^{\mathrm{task}}}$ and human demonstrations $\src{\mathcal D_H^{\mathrm{task}}}$ collected for each task and used in post-training.}
    \label{tab:tasks}
    \small
    \setlength{\tabcolsep}{2.5pt}
    \begin{tabular}{llccrr}
        \toprule
        Task & Data & Actions & Lang. & Traj. & Hours \\
        \midrule
        \multirow{2}{*}{\makecell[l]{Gear\\assembly}} & $\tgt{\mathcal D_R^{\mathrm{task}}}$ & \checkmark & $\times$ & 1{,}466 & 9.0 \\
        & $\src{\mathcal D_H^{\mathrm{task}}}$ & $\times$ & $\times$ & 2{,}715 & 12.0 \\
        \midrule
        \multirow{2}{*}{\makecell[l]{Bimanual\\pouring}} & $\tgt{\mathcal D_R^{\mathrm{task}}}$ & \checkmark & $\times$ & 474 & 2.2 \\
        & $\src{\mathcal D_H^{\mathrm{task}}}$ & $\times$ & $\times$ & 2{,}466 & 13.4 \\
        \midrule
        \multirow{2}{*}{\makecell[l]{Flower\\insertion}} & $\tgt{\mathcal D_R^{\mathrm{task}}}$ & \checkmark & $\times$ & 793 & 7.3 \\
        & $\src{\mathcal D_H^{\mathrm{task}}}$ & $\times$ & $\times$ & 2{,}117 & 10.6 \\
        \midrule
        \multirow{2}{*}{\makecell[l]{Spoon\\scooping}} & $\tgt{\mathcal D_R^{\mathrm{task}}}$ & \checkmark & $\times$ & 491 & 4.0 \\
        & $\src{\mathcal D_H^{\mathrm{task}}}$ & $\times$ & $\times$ & 1{,}973 & 7.9 \\
        \bottomrule
    \end{tabular}
    \end{minipage}}%
    \usebox{\sidetallbox}\hfill
    \begin{minipage}[t]{0.570\textwidth}
    \centering
    \caption{{\textbf{Real-world success rates.} Each method is evaluated with 20 trials per task; we report the number of successful trials on each task and the average success rate (\%).} {All methods are post-trained on the same teleoperated demonstrations $\tgt{\mathcal D_R^{\mathrm{task}}}$; \method and OpenWAM-$\alpha$ additionally use the human demonstrations $\src{\mathcal D_H^{\mathrm{task}}}$, and \method is pre-trained on 2{,}678 hours of human video $\src{\mathcal D_H}$.} The vanilla VLA shares the architecture of \method but removes latent actions. Task descriptions and success criteria are given in \cref{sec:supp_real_tasks}, and the demonstrations of each task are listed in \cref{tab:tasks}.}
    \label{tab:real}
    \vfill
    \small
    \setlength{\tabcolsep}{2pt}
    \renewcommand{\arraystretch}{1.4}
    \begin{tabular}{l*{5}{c}}
        \toprule
        Method & \makecell{Gear\\Assembly} & \makecell{Bimanual\\Pouring} & \makecell{Flower\\Insertion} & \makecell{Spoon\\Scooping} & Avg.\ $\uparrow$ \\
        \midrule
        GR00T N1.7 & {9/20} & {10/20} & {6/20} & {8/20} & 41.25 \\
        {OpenWAM-$\alpha$} & {12/20} & {15/20} & {8/20} & {16/20} & {63.75} \\
        Vanilla VLA & {8/20} & {5/20} & {4/20} & {5/20} & 27.50 \\
        \rowcolor{oursrow} \method & {15/20} & {16/20} & {11/20} & {16/20} & 72.50 \\
        \bottomrule
    \end{tabular}\par\kern0pt
    \end{minipage}
\end{table*}

\begin{figure*}[!t]
    \centering
    \includegraphics[width=\textwidth]{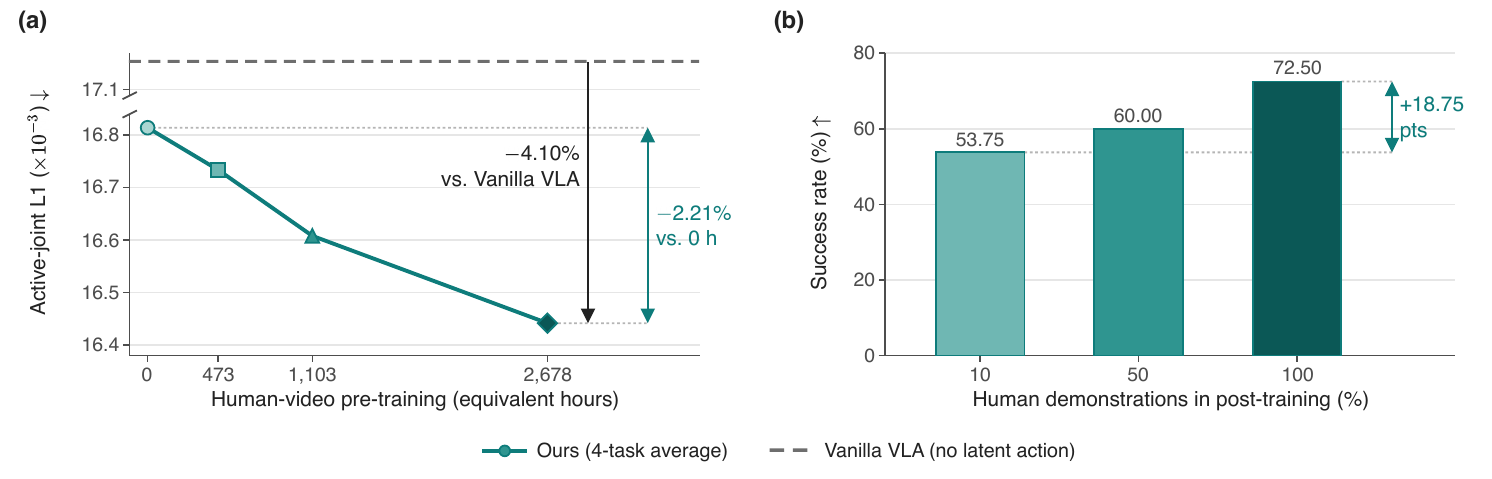}
    \caption{\textbf{Scaling with human video.} Both panels average over four tasks. \textbf{(a)}~Pre-training: validation L1 error of predicted actions on the joints each task uses, versus the pre-training budget in equivalent hours of human video. Gray dashed: vanilla VLA. \textbf{(b)}~Post-training: success rate versus the fraction of each task's human demonstrations $\src{\mathcal D_H^{\mathrm{task}}}$ used in post-training, starting from the model pre-trained on 2{,}678 hours of human video, with the teleoperated demonstrations $\tgt{\mathcal D_R^{\mathrm{task}}}$ fixed.}
    \label{fig:scaling}
\end{figure*}

\paragraph{Main results.}
We compare \method with three baselines, all post-trained on the same teleoperated demonstrations $\tgt{\mathcal D_R^{\mathrm{task}}}$.
GR00T N1.7~\citep{bjorck2025gr00t} and OpenWAM-$\alpha$~\citep{wang2026openwam} are open models that have undergone large-scale pre-training.
The vanilla VLA shares our architecture but removes latent actions, so it is trained on teleoperation data only and cannot learn from human videos; at each stage, it uses exactly the same teleoperation data as \method, $\tgt{\mathcal D_R}$ in mid-training and $\tgt{\mathcal D_R^{\mathrm{task}}}$ in post-training.
OpenWAM-$\alpha$ and \method additionally use the human demonstrations $\src{\mathcal D_H^{\mathrm{task}}}$ in post-training.
As shown in \cref{tab:real}, \method achieves the highest average success rate of 72.5\%, outperforming or matching every baseline on each task.
Compared with the strongest baseline OpenWAM-$\alpha$, \method requires a substantially lower training cost yet achieves a higher average success rate.
Since the vanilla VLA differs from \method only in latent actions, the gap between them measures what latent actions and the human videos they enable contribute.
We next examine the two stages at which human videos enter, pre-training and post-training, in \cref{fig:scaling}.

\Needspace{5\baselineskip}
\paragraph{Scaling in pre-training.}
Human videos provide an additional axis for scaling policy training beyond robot demonstrations.
We pre-train the policy with budgets equivalent to 0, 473, 1{,}103, and 2{,}678 hours of human video, and then mid-train each for one epoch over the teleoperation data $\tgt{\mathcal D_R}$ and post-train it on each task for 30{,}000 iterations.
We report the validation L1 error between predicted and ground-truth actions on held-out teleoperated demonstrations of the four tasks, counting only the joints each task uses.
We also compare with the vanilla VLA, with all other settings unchanged.
As shown in \cref{fig:scaling}(a), the validation L1 error decreases with every increase in the pre-training budget.
This improvement requires no additional robot data, since all models share the same teleoperation data in mid- and post-training.

Latent actions help even without human-video pre-training: our 0\,h variant already achieves a lower final validation L1 error than the vanilla VLA.
Because the latent actions are grounded in robot control, predicting them shapes the representations from which the action expert generates actions, and they let the human demonstrations of each task supervise the same action expert, which the vanilla VLA cannot exploit.
With 2{,}678 hours of human-video pre-training, the error drops further below both the vanilla VLA and our 0\,h variant, with the relative reductions marked in \cref{fig:scaling}(a).

\Needspace{5\baselineskip}
\paragraph{Scaling in post-training.}
At the task level, starting from the model pre-trained on 2{,}678 hours of human video, we post-train each of the four tasks with 10\%, 50\%, and 100\% of its human demonstrations $\src{\mathcal D_H^{\mathrm{task}}}$ while keeping the teleoperated demonstrations $\tgt{\mathcal D_R^{\mathrm{task}}}$ fixed.
As shown in \cref{fig:scaling}(b), the average success rate increases steadily with more human demonstrations, from 53.75\% to 72.5\%.
Since human demonstrations require no robot to record, collecting more of them for a task offers an inexpensive way to improve the policy on that task, complementary to teleoperation.

\subsection{Ablations and Analysis}
\label{sec:ablation}

We examine the robot-anchored semantic transfer described in \cref{sec:semantic_transfer} in two steps.
We first ablate the alignment loss and analyze what it contributes to the latent space, within a single embodiment and across embodiments; we then ablate two designs that determine how alignment acts, the stop-gradient that keeps robot latents as control anchors and the SoftVQ quantizer.

\begin{figure*}[!t]
    \centering
    \includegraphics[width=1\textwidth]{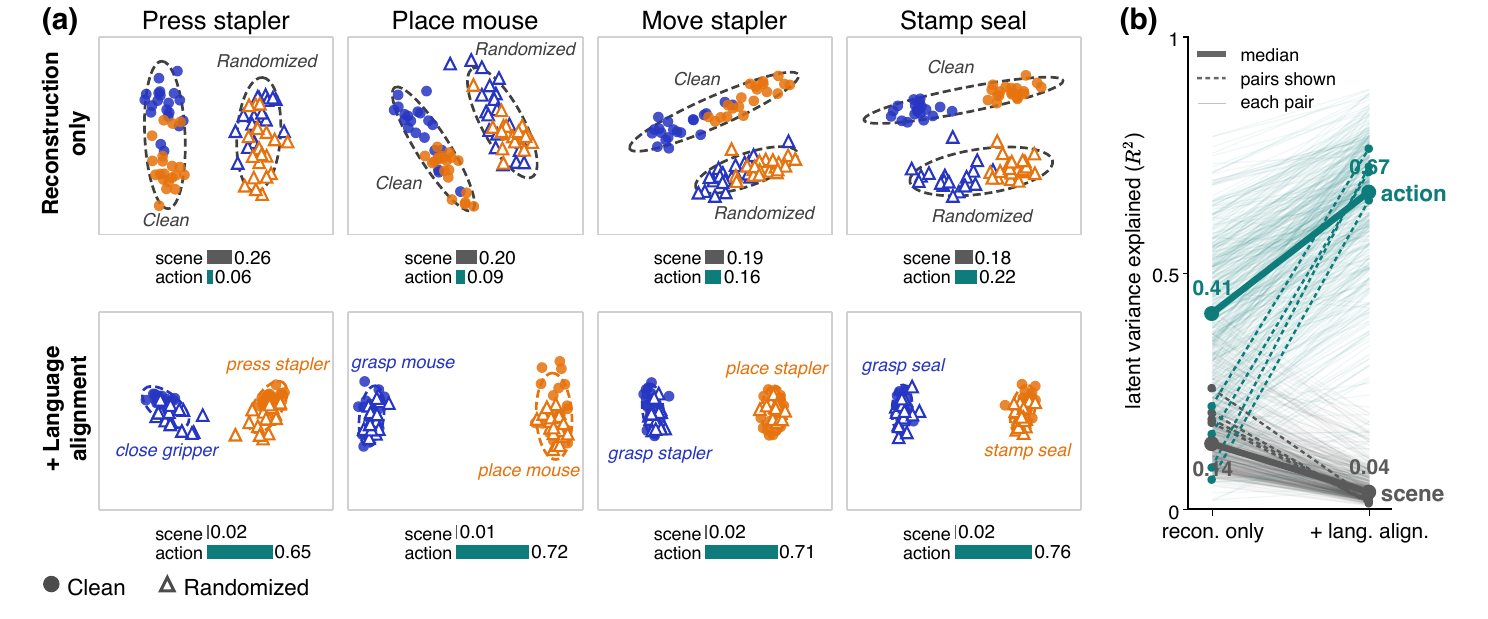}
    \caption{\textbf{Latent actions without and with language alignment.} \textbf{(a)} For four RoboTwin 2.0 tasks, \src{ARX-X5} latents $\mathbf z$ of two atomic actions of the right arm, with 20 clips per action and scene type, projected to 2D by PCA fitted to each panel. Color: action; marker: scene type. Bars report the fraction of variance in the full latent space explained by scene and by action. \textbf{(b)} These fractions across all analyzed pairs: thick lines show medians; dashed lines mark the pairs in (a). Alignment shifts the organization from scene toward action.}
    \label{fig:scene_pairs}
\end{figure*}

\Needspace{16\baselineskip}
\begin{wraptable}{r}{0.42\textwidth}
    \centering
    \vspace{-12pt}
    \caption{\textbf{Tokenizer ablations on RoboTwin 2.0.} Each variant changes one design choice of the full tokenizer, and the policy is trained in the same way. w/o $\operatorname{sg}$: bidirectional alignment. Success rates (\%) on \tgt{Aloha-AgileX}.}
    \label{tab:abl_sim}
    \small
    \begin{tabular}{lcc}
        \toprule
        Tokenizer & Clean & Random \\
        \midrule
        \rowcolor{oursrow} \method (full) & 80.2 & 48.6 \\
        \midrule
        \multicolumn{3}{l}{\textit{Semantic alignment}} \\
        w/o $\mathcal L_{\mathrm{align}}$ & 75.5 & 36.8 \\
        \midrule
        \multicolumn{3}{l}{\textit{Stop-gradient}} \\
        w/o $\operatorname{sg}$ & 73.5 & 46.8 \\
        \bottomrule
    \end{tabular}
    \vspace{-8pt}
\end{wraptable}

\paragraph{Semantic alignment.}
On RoboTwin 2.0, we test whether the correspondence between latent actions of the two embodiments requires explicit supervision or emerges from reconstruction alone.
Since reconstruction constrains each latent only through its own clip, $\mathcal L_{\mathrm{align}}$ is the only term in \cref{eq:tokenizer_objective} that relates the two embodiments; we set $\lambda_{\mathrm{align}}=0$ and keep everything else unchanged.
As shown in \cref{tab:abl_sim}, removing alignment lowers success rates in both settings, so reconstruction alone does not provide this correspondence.
The drop is larger in Random, where scenes vary.
\Cref{fig:scene_pairs} examines the \src{ARX-X5} latents, which can be organized by scene as well as by action since the videos are recorded in both clean and randomized scenes.
Without alignment, the clips of three of the four tasks in \cref{fig:scene_pairs}(a) split by scene, with the two actions mixed in each group; with alignment, they split by atomic action, since clips with semantically similar descriptions receive similar alignment targets regardless of the scene.
As shown in \cref{fig:scene_pairs}(b), across all analyzed pairs, alignment raises the median fraction of latent variance explained by the action from 0.41 to 0.67 and lowers that explained by the scene from 0.14 to 0.04.

\begin{figure*}[!t]
    \centering
    \includegraphics[width=\textwidth]{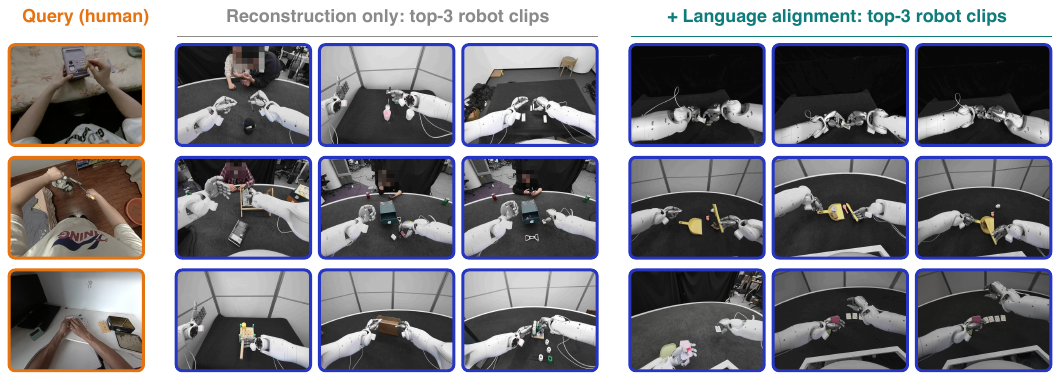}
    \caption{\textbf{Cross-embodiment retrieval for human pre-training videos.} Each human query (left) and its three nearest robot clips over all tasks, under the reconstruction-only tokenizer (middle) and \method (right).}
    \label{fig:retrieval}
\end{figure*}

\paragraph{Cross-embodiment retrieval.}
\Cref{fig:scene_pairs} concerns clips of a single embodiment; we next test whether alignment also relates clips across embodiments.
Given a query clip of a non-target embodiment, we retrieve the target-robot clips whose latent actions are most similar to it, searching over all tasks with the similarity used by the alignment loss.
\Cref{fig:retrieval} shows human queries from the pre-training videos $\src{\mathcal D_H}$, of tasks absent from the robot data.
Human and robot videos differ in viewpoint, embodiment, and background, and the reconstruction-only tokenizer returns robot clips unrelated to the query.
\method instead returns semantically related robot behaviors; for example, a human sweeping the floor retrieves the robot sweeping trash into a dustpan.
\Cref{sec:supp_retrieval} evaluates retrieval quantitatively on RoboTwin 2.0 and in the real world.

\Needspace{16\baselineskip}
\begin{wrapfigure}{r}{0.42\textwidth}
    \centering
    \vspace{-12pt}
    \includegraphics[width=\linewidth]{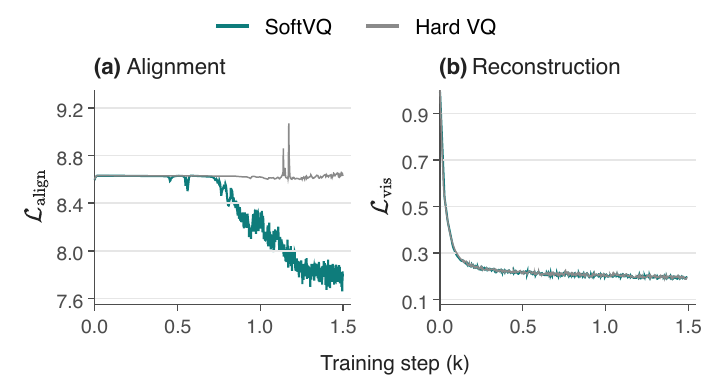}
    \caption{\textbf{Tokenizer training with SoftVQ and hard VQ.} SoftVQ reduces (a)~the alignment loss, whereas hard VQ leaves it nearly unchanged, although both reach comparable (b)~visual reconstruction losses. All other settings are identical.}
    \label{fig:quantizer_losses}
    \vspace{-10pt}
\end{wrapfigure}

\paragraph{Stop-gradient.}
The stop-gradient makes alignment one-way: robot latents serve as fixed targets, and only human latents move toward them.
Removing $\operatorname{sg}$ from $s^{z}_{ij}$ in \cref{eq:semantic_alignment} makes alignment bidirectional, so robot latents are also pulled toward human latents, which lack action supervision.
Robot latents then have to serve two objectives, encoding the robot's own actions and matching human latents, and the second can compromise the first.
As shown in \cref{tab:abl_sim}, bidirectional alignment lowers success rates, most notably in the clean setting, indicating that robot latents should remain anchored to control.

\paragraph{Quantizer.}
To test the choice of quantizer in \cref{sec:semantic_transfer}, we replace SoftVQ with a standard hard VQ.
\Cref{fig:quantizer_losses} shows that SoftVQ substantially reduces the alignment loss, whereas hard VQ keeps it near the value of a uniform prediction, even though both reach comparable visual reconstruction losses.
Since the alignment loss is computed on the pre-quantization latents, hard VQ hinders alignment already before quantization, consistent with the argument in \cref{sec:semantic_transfer}.

%% file: sections/3_related.tex
\section{Related Work}
\label{sec:related}

\paragraph{Vision-language-action models.}
VLA models build on pre-trained vision-language models to map visual observations and language instructions to robot actions~\citep{brohan2023rt,kim2024openvla,black2024pi_0,intelligence2025pi05,bjorck2025gr00t,zheng2025xvla,yuan_qwen-robotmanip_2026}.
Their generalization depends on the diversity of robot data, which is costly to collect through teleoperation.
Language has also been used to structure action learning in VLAs, through interpretable action primitives~\citep{weng_language-grounded_2026} or language-conditioned action priors learned without visual input~\citep{lin_la4vla_2026}.
\method keeps the VLA design, with a flow-matching action expert, and adds latent-action prediction so that videos without action labels can supervise the same action expert.

\paragraph{Learning from human videos.}
Human videos have been used to pre-train visual representations for robot manipulation~\citep{nair2022r3m}.
Closer to policy learning, one line of work extracts explicit motion from human videos, such as hand trajectories or poses, and maps it to robot actions~\citep{wang2023mimicplay,kareer2024egomimic,qiu2025humanoid,yang2025egovla,luo2025beingh0,fu_metis_2025}; this requires hand tracking and a correspondence between human and robot kinematics.

With sufficiently diverse pre-training, co-training VLAs on such human data can yield human-to-robot transfer~\citep{kareer_emergence_2025,zheng2026egoscale}.
Another line predicts future observations as a proxy for actions: generating video plans that are then converted into actions~\citep{du2023unipi,ko2023avdc}, predicting subgoal images~\citep{zhao_cot-vla_2025,yang_world-language-action_2026}, or training WAMs that predict videos jointly with actions~\citep{wu2023gr1,yuan2026fastwam,li2026lingbotva,zhang_native_2026,nvidia_cosmos_2026,wang2026openwam}.
Other work predicts future representations rather than pixels~\citep{su_world_2026,miao_jepa-vla_2026,tang_towards_2026,ma2026internvlaa15}; when co-training on egocentric human videos, feature and 3D-flow targets transfer to robots better than pixel targets~\citep{li_egowam_2026}.
Video prediction requires modeling scene appearance in addition to motion, and in \cref{sec:sim} some WAMs degrade sharply under domain randomization.
\method instead supervises the policy with latent actions, which require neither hand tracking nor pixel-level prediction.

\paragraph{Latent action models.}
Instead of labeling videos with an inverse dynamics model trained on action-labeled data~\citep{baker_video_2022}, latent action models infer actions from consecutive frames without action labels, by learning latents that predict the next frame~\citep{schmidt2023learning,bruce2024genie,chen_igor_2024,liang_clam_2026}.
They have been used to pre-train policies on actionless videos, including human videos~\citep{ye2025latent,chen2024moto,bu2025univla,chen2026iclr-villax,li2026rotvla}.
Because the latents are learned by reconstructing each clip, they can capture scene changes and distractors rather than the action~\citep{NEURIPS2025_d7b3cef7}.
Prior work addresses this by conditioning on task instructions~\citep{bu2025univla}, by structuring the latent space~\citep{li2026rotvla}, by contrastive disentanglement with action-category priors~\citep{dai_conla_2026}, or by adversarially suppressing embodiment- and environment-specific factors~\citep{liu_scar_2026}.
To relate embodiments, other methods align latents with robot actions recorded alongside the videos~\citep{zhang_clap_2026,fan_xr-1_2026}, with action labels of human hands~\citep{luo_joint-aligned_2026}, through paired human-robot demonstrations~\citep{zhu_harp-vla_2026}, or through the shared visual consequences of heterogeneous kinematics~\citep{chen2026unit}; a related approach contrastively aligns the action spaces of human hands, dexterous hands, and grippers~\citep{bauer_latent_2026}.
\method instead uses atomic-action descriptions to relate unpaired human and robot clips by graded semantic similarity, and aligns human latents to robot latents grounded in control, so that human and robot executions of semantically similar actions receive similar latents.

%% file: sections/4_conclusion.tex
\section{Conclusion}
\label{sec:conclusion}
We presented \method, which uses language to align human and robot latent actions for policy learning.
Natural-language descriptions of atomic actions relate unpaired human and robot clips through graded semantic similarity, and \method turns these relations into soft neighborhood targets that move human latents toward robot latents grounded in control.
The resulting latent actions let both general human videos and human demonstrations of the target task supervise the same policy that controls the robot.
Experiments show that this supervision is scalable and effective.

Our study has limitations that point to future work.
The tokenizer requires atomic-action descriptions for its training clips, although the policy itself learns from video frames alone.
Our pre-training scaling analysis covers up to 2{,}678 hours; whether the trend holds at larger scales remains open.

%% file: sections/X_suppl.tex
\section{Cross-Embodiment Retrieval}
\label{sec:supp_retrieval}

\paragraph{Protocol.}
As in \cref{sec:ablation}, each query clip of a non-target embodiment retrieves the most similar target-robot clip over all tasks, using the similarity of the alignment loss, \ie, the cosine similarity between pre-quantization latents $\mathbf z$ averaged over tokens.
Besides the reconstruction-only tokenizer and \method, we report two references: \emph{Random}, a uniformly sampled robot clip, and \emph{Language}, which retrieves by the similarity of atomic-action descriptions.
On RoboTwin 2.0, \src{ARX-X5} clips from 47 tasks query all \tgt{Aloha-AgileX} clips; we check whether the retrieved clip performs the same atomic action and measure the distance between the end-effector displacements of the two clips.
In the real world, \src{human} clips query all \tgt{robot} teleoperation clips.
For queries from the 56 tasks shared with the robot data, we check whether the retrieved clip comes from the same task and whether both hands perform the same verbs.
For these queries and for those from tasks absent from the robot data, we also compute the similarity between the atomic-action descriptions of the query and the retrieved clip, rescaled so that Random scores 0 and Language scores 1.

\begin{table}[!htb]
    \centering
    \caption{\textbf{Top-1 cross-embodiment retrieval over all tasks.} Language retrieval is omitted on RoboTwin 2.0, where the scripted descriptions of the same atomic action are identical. Language similarity is rescaled so that Random scores 0 and Language scores 1.}
    \label{tab:retrieval}
    \small
    \setlength{\tabcolsep}{4pt}
    \begin{tabular}{lcccc}
        \toprule
        Metric & Random & Recon.\ only & + Lang.\ align. & Language \\
        \midrule
        \multicolumn{5}{l}{\textit{RoboTwin 2.0: \src{ARX-X5} $\rightarrow$ \tgt{Aloha-AgileX}}} \\
        Same atomic action (\%) $\uparrow$ & 0.8 & 41.3 & \textbf{78.2} & -- \\
        End-effector displacement error (cm) $\downarrow$ & 8.03 & 2.89 & \textbf{2.58} & -- \\
        \midrule
        \multicolumn{5}{l}{\textit{Real world: \src{human} $\rightarrow$ \tgt{robot}, shared tasks}} \\
        Same task (\%) $\uparrow$ & 1.7 & 3.8 & \textbf{34.2} & 55.0 \\
        Same verbs of both hands (\%) $\uparrow$ & 4.1 & 6.2 & \textbf{17.0} & 66.4 \\
        Language similarity $\uparrow$ & 0.00 & 0.04 & \textbf{0.32} & 1.00 \\
        \midrule
        \multicolumn{5}{l}{\textit{Real world: \src{human} $\rightarrow$ \tgt{robot}, tasks absent from robot data}} \\
        Language similarity $\uparrow$ & 0.00 & 0.06 & \textbf{0.31} & 1.00 \\
        \bottomrule
    \end{tabular}
\end{table}

\begin{figure*}[tp]
    \centering
    \begin{tabular}{@{}l@{\hspace{4pt}}c@{}}
        \raisebox{0.49\textwidth}{\textbf{(a)}} & \includegraphics[width=0.96\textwidth]{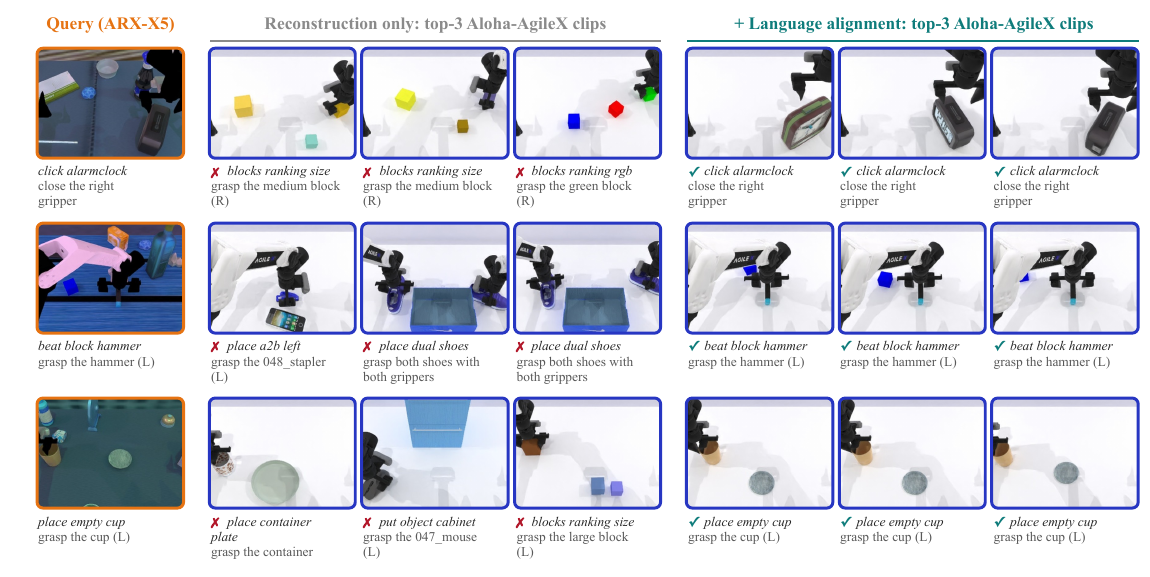} \\[2pt]
        \raisebox{0.49\textwidth}{\textbf{(b)}} & \includegraphics[width=0.96\textwidth]{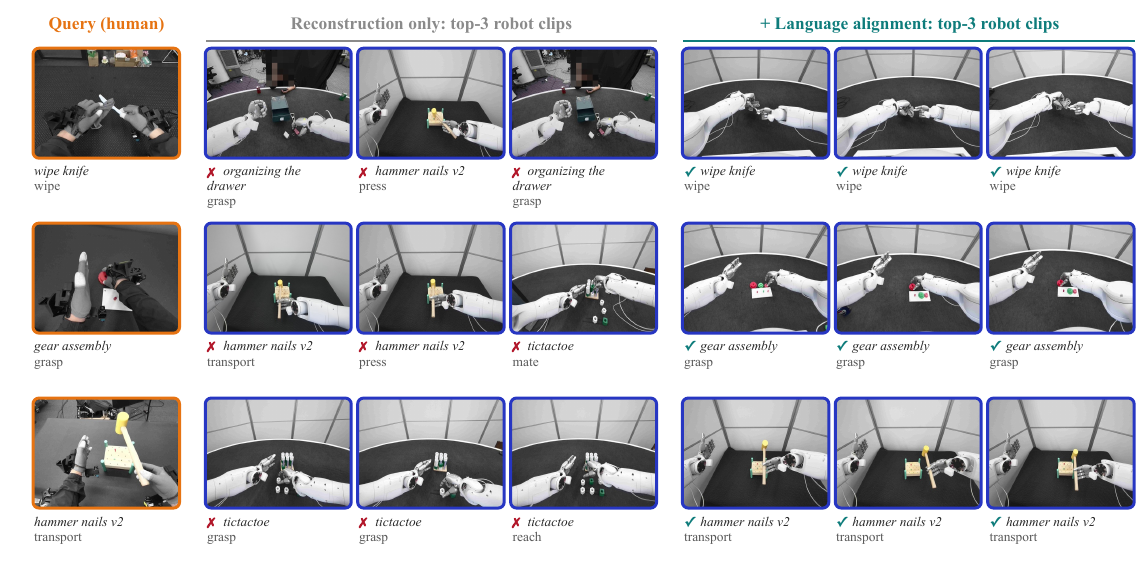} \\
    \end{tabular}
    \caption{\textbf{Cross-embodiment retrieval on shared tasks.} Each query (left) and its three nearest target-robot clips under the reconstruction-only tokenizer (middle) and \method (right). \textbf{(a)}~RoboTwin 2.0: \src{ARX-X5} queries, \tgt{Aloha-AgileX} clips. \textbf{(b)}~Real world: \src{human} queries, \tgt{robot} clips. Each clip is labeled with its task (italic) and atomic action; \checkmark/$\times$ mark whether the atomic action (a) or the task and verb (b) match the query. Examples are selected.}
    \label{fig:retrieval_paired}
\end{figure*}

\begin{figure*}[tp]
    \centering
    \includegraphics[width=0.95\textwidth]{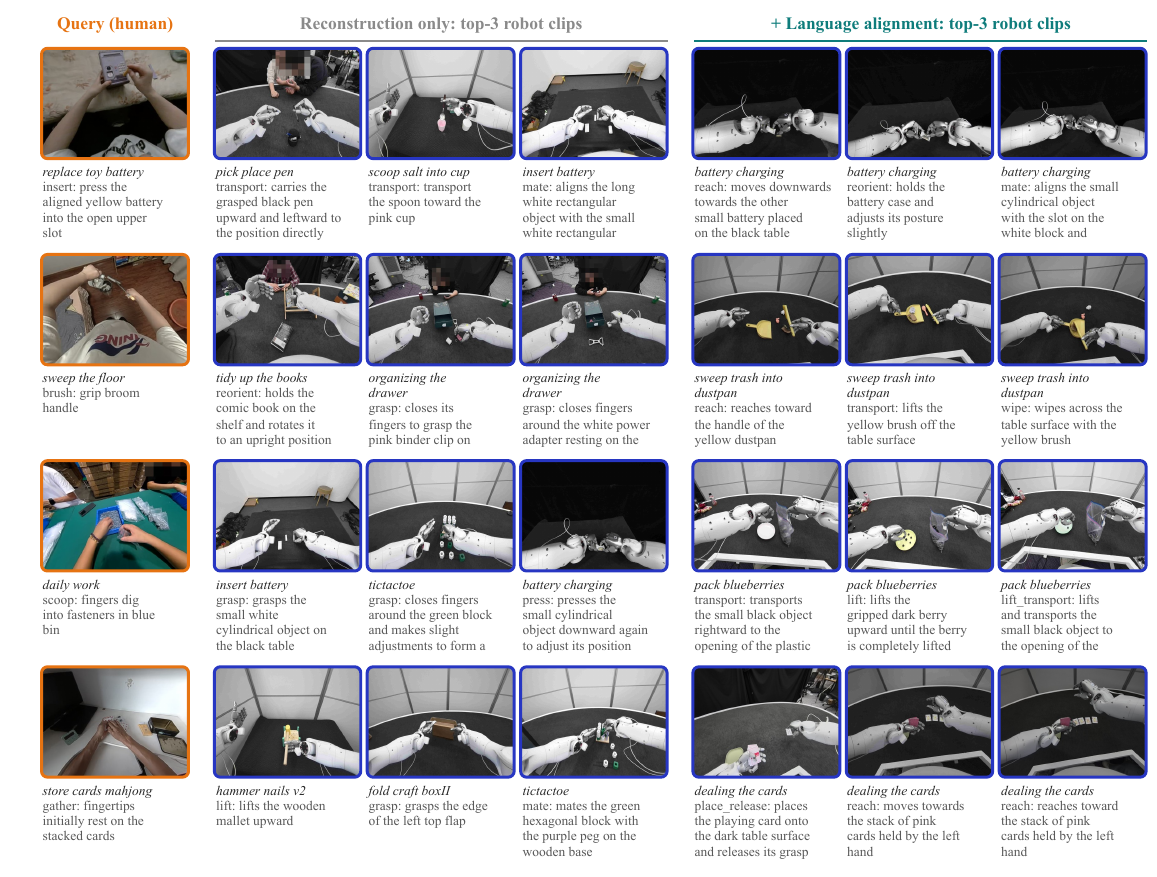}
    \caption{\textbf{Retrieval for \src{human} clips of tasks absent from the \tgt{robot} data.} Annotated version of \cref{fig:retrieval} with one additional query.}
    \label{fig:retrieval_unpaired}
\end{figure*}

\paragraph{Results.}
Alignment improves every metric in \cref{tab:retrieval}.
On RoboTwin 2.0, the two embodiments share the viewpoint and workspace, so reconstruction alone already retrieves the same atomic action for 41.3\% of queries; alignment raises this to 78.2\%, with gains in all 47 tasks.
In the real world, human and robot videos differ in viewpoint, embodiment, and background, and reconstruction alone is close to random.
Alignment raises the same-task rate from 3.8\% to 34.2\%, and its language similarity on tasks absent from the robot data (0.31) is close to that on shared tasks (0.32).
\Cref{fig:retrieval_paired} shows examples on shared tasks and \cref{fig:retrieval_unpaired} on tasks absent from the robot data; for instance, a human sweeping the floor with a broom retrieves the robot sweeping trash into a dustpan.

\section{Latent-Space Analysis of Semantic Alignment}
\label{sec:supp_latent_analysis}
\begin{figure*}[tp]
    \centering
    \includegraphics[width=\textwidth]{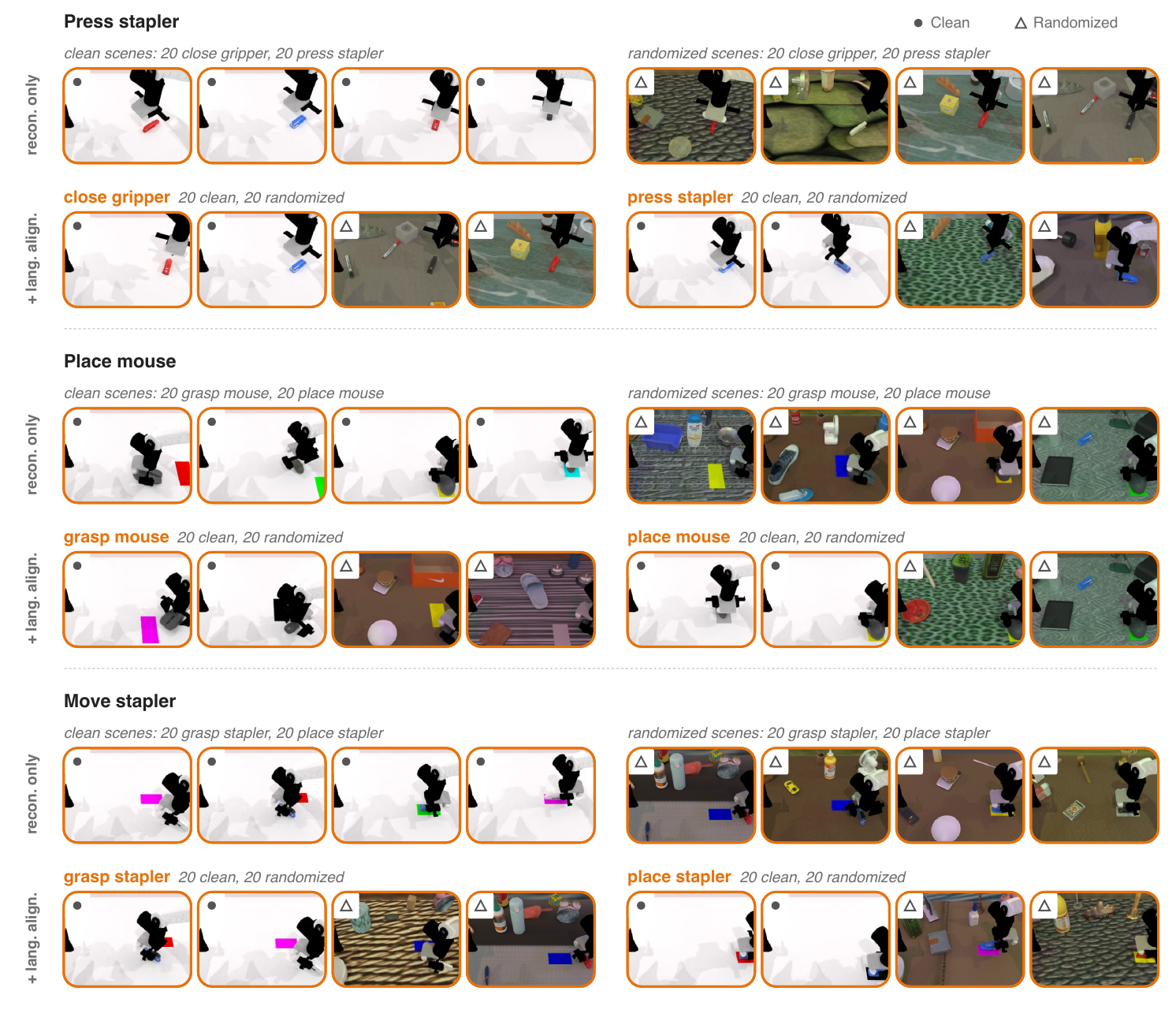}
    \caption{\textbf{Clusters of latent actions.} For three pairs of atomic actions in \cref{fig:scene_pairs}(a), k-means ($k=2$) is applied to the 80 \src{ARX-X5} latents of each tokenizer, 20 per atomic action and scene type. Each half-row shows one cluster: its composition in the header and four of its clips. Without alignment (recon.\ only), the two clusters are the clean and the randomized clips, each mixing both atomic actions; with alignment, each cluster is one atomic action, covering both scene types. Filled circle: clean scene; open triangle: randomized scene.}
    \label{fig:cluster_gallery}
\end{figure*}

\Cref{fig:scene_pairs}(a) shows that language alignment reorganizes \src{ARX-X5} latents by atomic action rather than by scene type, \ie, clean or randomized.
\Cref{fig:cluster_gallery} illustrates this change by clustering the latents of three pairs in \cref{fig:scene_pairs}(a) into two groups: without alignment, the two clusters are exactly the 40 clean and the 40 randomized clips, whereas with alignment, each cluster contains exactly one atomic action with 20 clips from each scene type.

\section{Implementation Details}
\label{sec:supp_impl}

\paragraph{Tokenizer.}
\Cref{tab:supp_tokenizer} lists the tokenizer settings of both experiments.
Each observation is resized to $224\times224$ and encoded by a frozen DINOv2-L, using features from its second-to-last layer.
The visual encoder, a transformer with 8 layers and 16 heads, maps the endpoint features of a clip to 8 latent tokens.
The quantizer is SoftVQ with 256 codes and a softmax temperature of 0.25.
The visual decoder, also with 8 layers, predicts the DINOv2 features of the future observation, and the action decoder predicts the action chunk from the quantized latents and the initial robot state with a Huber loss ($\delta=0.35$).
The visual reconstruction loss has weight 1, and the softmax temperature of $Q_i$ in \cref{eq:semantic_alignment} is 0.2.
To provide robot clips as alignment targets for every human clip, each batch draws human and robot clips in equal numbers, and robot clips are gathered across GPUs.
We optimize with AdamW (learning rate $10^{-4}$, $\beta=(0.9,0.95)$, weight decay 0.01) at a constant learning rate, with gradient clipping at 1.0 and bf16 mixed precision.

\begin{table}[h]
    \centering
    \caption{\textbf{Tokenizer settings.} Horizon $k$: frame interval between the two endpoint observations of a clip. In the real world, $k$ is 8 frames for one source of human videos, and 16, 8, or 4 frames for the other, chosen by motion speed so that slow motions span longer intervals.}
    \label{tab:supp_tokenizer}
    \small
    \begin{tabular}{lcc}
        \toprule
        & Simulation & Real world \\
        \midrule
        Camera views & 1 (head) & 3 (head, two wrists) \\
        Hidden size & 768 & 1024 \\
        Codebook dimension & 32 & 64 \\
        Action decoder layers & 8 & 4 \\
        Text encoder & CLIP ViT-B/32~\citep{radford2021clip} & Qwen3-Embedding-4B \\
        Temperature of $P_i$ & 0.0253 & 0.049 \\
        $\lambda_{\mathrm{act}}$ & 10 & 5 \\
        $\lambda_{\mathrm{align}}$ & 0.001 & 0.005 \\
        \multirow{2}{*}{Horizon $k$ (frames)} & \tgt{Aloha-AgileX}: 8 & robot: 16 \\
        & \src{ARX-X5}: 6 & human: 4--16 \\
        \bottomrule
    \end{tabular}
\end{table}

\paragraph{Policy.}
The VLM is Qwen3-VL-4B, and the action expert is a separately parameterized Qwen3-style transformer with 0.6B parameters (36 layers, hidden size 768) that attends jointly with the VLM at every layer.
The latent-action loss weight in \cref{eq:policy_objective} is $\lambda_{\mathrm{LA}}=0.1$.
Pre-training starts from the Qwen3-VL-4B-Instruct weights with a randomly initialized action expert and 8 latent queries; mid-training starts from the pre-trained checkpoint, and post-training from the mid-trained one, keeping the same architecture throughout.

\begin{figure}[h]
    \centering
    \includegraphics[width=\linewidth]{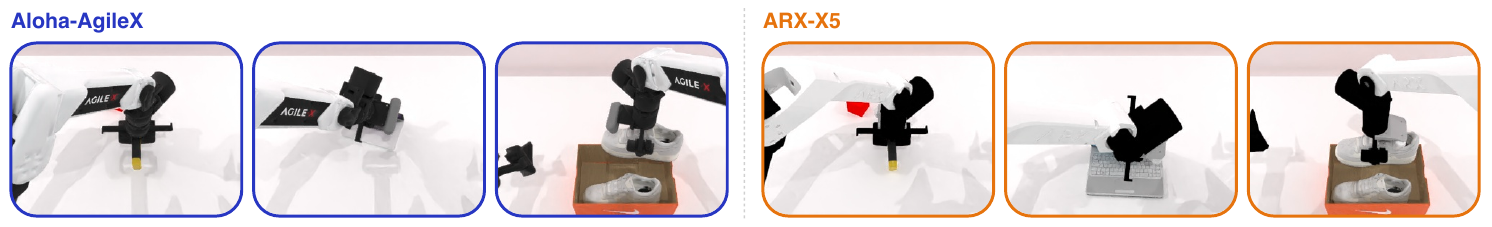}
    \caption{\textbf{Embodiments in RoboTwin 2.0.} Frames of the same task performed by the target embodiment \tgt{Aloha-AgileX} (left), whose complete trajectories form $\tgt{\mathcal D_R}$, and the non-target embodiment \src{ARX-X5} (right), whose videos form $\src{\mathcal D_H}$.}
    \label{fig:robotwin_embodiments}
\end{figure}

\section{Atomic-Action Annotations on RoboTwin 2.0}
\label{sec:supp_robotwin_annotation}
\begin{figure}[h]
    \centering
    \includegraphics[width=\linewidth]{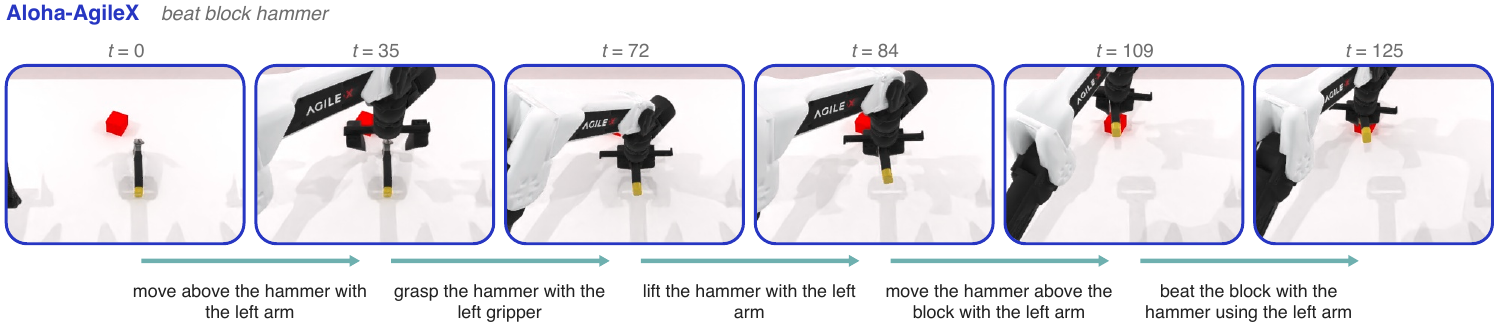}
    \caption{\textbf{Atomic-action annotations on RoboTwin 2.0.} Frames of one \tgt{Aloha-AgileX} episode of the \emph{beat block hammer} task at the boundaries between consecutive steps of the task program. The description between two frames is the atomic action performed between them.}
    \label{fig:supp_robotwin_annotation}
\end{figure}

RoboTwin 2.0 collects expert trajectories by running a scripted program for each task with a motion planner~\citep{chen2025robotwin}.
The program composes the task from manipulation primitives, and the planner executes each primitive as a short sequence of arm motions and gripper commands.
This execution divides each trajectory into atomic actions whose boundaries are given by the program.
The program also specifies the target object and the executing arm of each atomic action, so we describe each atomic action in language without manual annotation.
These atomic actions typically last about one second.
\Cref{fig:supp_robotwin_annotation} shows one episode divided in this way.
We annotate the \src{ARX-X5} and \tgt{Aloha-AgileX} trajectories with the same procedure.

\section{Atomic-Action Annotations in the Real World}
\label{sec:supp_real_annotation}
\begin{figure}[h]
    \centering
    \includegraphics[width=\linewidth]{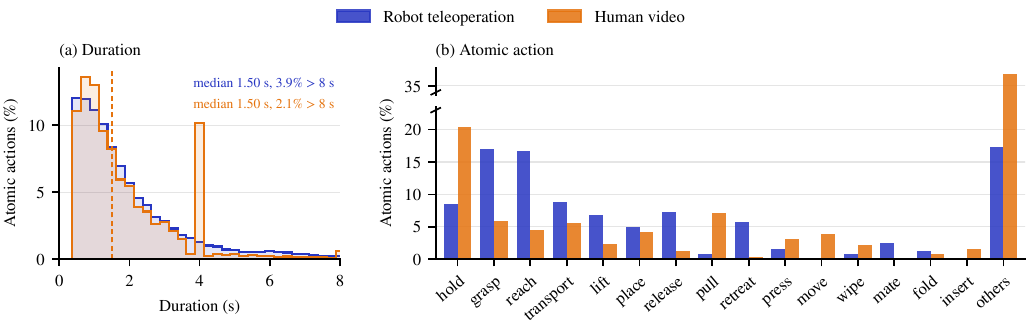}
    \caption{\textbf{Statistics of real-world atomic-action annotations.} \textbf{(a)}~Distribution of atomic-action durations, with dashed lines marking the medians. \textbf{(b)}~Frequency of the 15 most frequent predefined atomic actions, where others include composite labels and labels introduced by the VLM. Percentages are over the atomic actions of both hands.}
    \label{fig:supp_atomic_stats}
\end{figure}

Real-world trajectories are not generated by a task program, so we annotate their atomic actions from the head-camera video with a vision-language model (VLM), Doubao-Seed-2.0-Pro~\citep{bytedance2026seed2}.
Each trajectory is first divided into subtasks, which come from existing subtask annotations where available and are otherwise proposed by the VLM from the video sampled at 1\,Hz.
The VLM then divides each subtask, sampled at 4\,Hz, into the atomic actions of each hand.
An atomic action is a short, continuous interval with a single manipulation intent, and its boundaries lie at observable changes, such as the start or end of a motion or contact and separation between a hand and an object.
The VLM labels each atomic action with one of over a dozen predefined common actions, or with a label of its own when none of them applies, and describes it in one sentence that starts with ``The left hand'' or ``The right hand''.
\Cref{tab:supp_atomic_defs} defines the 15 most frequent actions.
Robot hands are referred to in the same way as human hands, so the descriptions do not depend on the embodiment.
\Cref{fig:supp_atomic_stats} shows the distributions of the durations and labels of atomic actions.
The peak at 4\,s comes from long hold intervals in human videos, which are annotated as consecutive 4\,s segments.
\begin{table}[h]
    \centering
    \caption{\textbf{Definitions of the 15 most frequent atomic actions.}}
    \label{tab:supp_atomic_defs}
    \small
    \begin{tabularx}{\linewidth}{@{}lX@{}}
        \toprule
        Action & Definition \\
        \midrule
        hold & The hand stably holds an object, with the hand, fingers, wrist, and object all still. \\
        grasp & The fingers close to form a stable grasp, including visible adjustments around it. \\
        reach & The empty hand moves toward a target and ends close to it before forming a grasp. \\
        transport & The hand carries an object that has left its support to a new location. \\
        lift & The hand raises a grasped object until it first fully leaves its support. \\
        place & The hand puts an object at a target location, where it becomes stably supported. \\
        release & The fingers open to end the grasp, and the hand separates from the object. \\
        pull & The hand pulls an object toward the body while keeping contact or a grasp. \\
        retreat & The empty hand withdraws from the manipulation position. \\
        press & The hand pushes on an object along the surface normal until it is visibly pressed in or triggered. \\
        move & The hand moves with a clear displacement that no more specific action describes. \\
        wipe & The hand or a held tool keeps contact with a surface and moves along it to wipe it. \\
        mate & The hand aligns two objects and brings them into an inserted, nested, or fitted relation. \\
        fold & The hand bends a deformable object so that one part covers or lies against another. \\
        insert & The hand pushes an aligned object through an opening until it is inside a hole, slot, or container. \\
        \bottomrule
    \end{tabularx}
\end{table}

\section{Real-World Platform}
\label{sec:supp_platform}
\begin{figure}[h]
    \centering
    \begin{tikzpicture}
        \node[anchor=south west, inner sep=0] (img) at (0,0) {\includegraphics[width=0.36\linewidth]{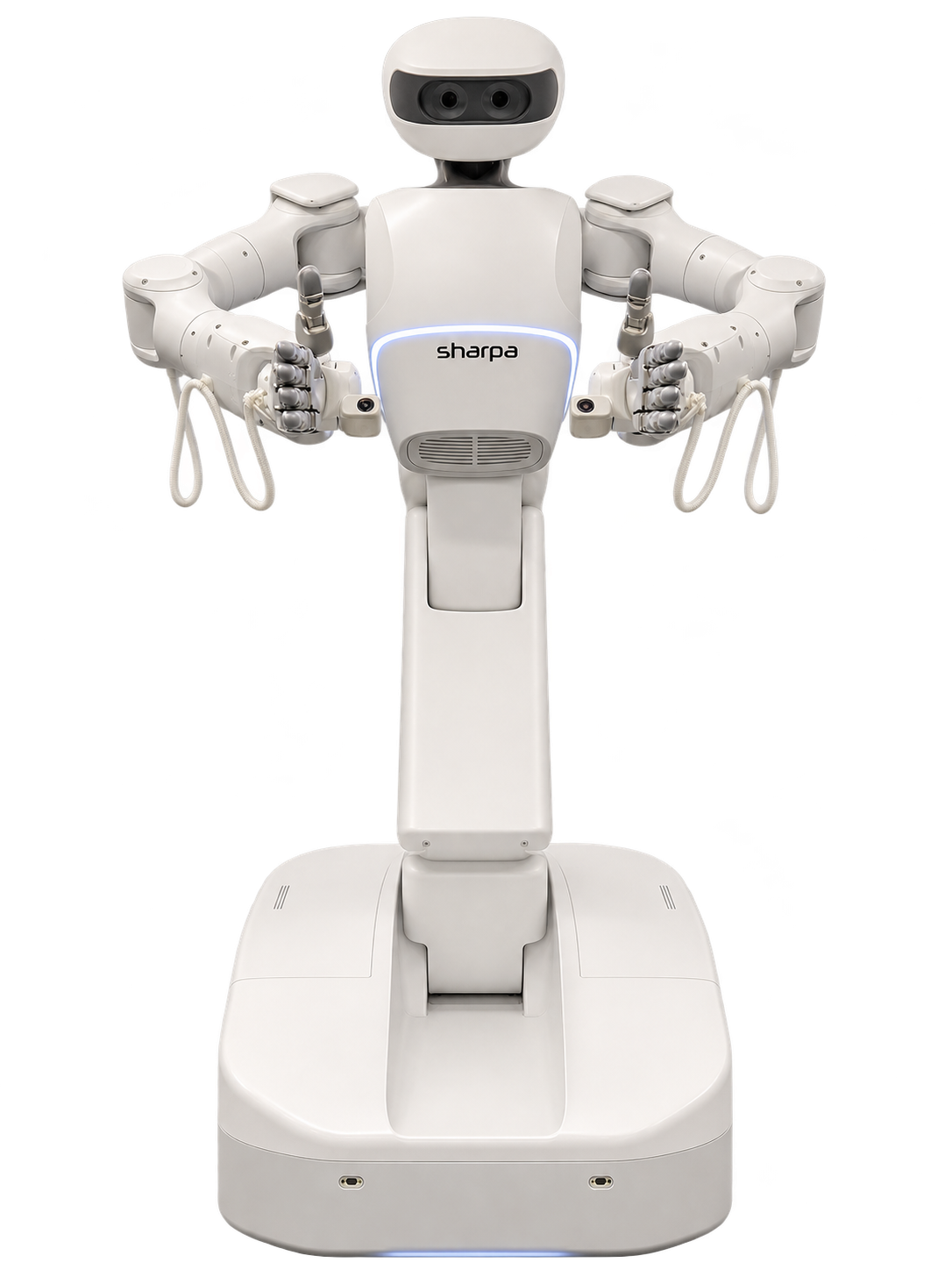}};
        \begin{scope}[x={(img.south east)}, y={(img.north west)}, font=\small, every path/.style={draw=black!70, line width=0.4pt}]
            \foreach \px/\py/\lx/\ly/\lab/\side in {
                0.50/0.925/-0.12/0.925/(a)/left,
                0.30/0.700/-0.12/0.700/(c)/left,
                0.47/0.490/-0.12/0.490/(e)/left,
                0.80/0.790/1.12/0.790/(b)/right,
                0.625/0.668/1.12/0.520/(d)/right,
                0.62/0.150/1.12/0.150/(f)/right}{
                \fill[black!70] (\px,\py) circle (1.2pt);
                \draw (\px,\py) -- (\lx,\ly) node[\side, inner sep=2pt] {\lab};
            }
        \end{scope}
    \end{tikzpicture}
    \caption{\textbf{Real-world platform.} The Sharpa North robot comprises (a) a head-mounted stereo camera, (b) two 7-DoF arms, (c) two 22-DoF Sharpa Wave dexterous hands, (d) two wrist-mounted fisheye cameras, (e) a torso, and (f) a wheeled base.}
    \label{fig:supp_platform}
\end{figure}

All real-world experiments use a Sharpa North robot\footnote{\url{https://www.sharpa.com/pages/north}}, a wheeled full-body humanoid with two 7-DoF arms, each equipped with a 22-DoF Sharpa Wave dexterous hand\footnote{\url{https://www.sharpa.com/pages/wave}}.
The Wave is a human-scale, five-finger hand with a vision-based tactile sensor in each fingertip; our policy does not use tactile inputs.
The policy observes RGB images from a head-mounted stereo camera and two wrist-mounted fisheye cameras, together with the joint angles of both arms, both hands, and five torso and neck joints.
It predicts chunks of 32 joint-angle targets at 60\,Hz, expressed relative to the current state and covering 63 actuated DoF, while images are captured at 30\,Hz.
\Cref{fig:supp_platform} shows the robot and its main components.

\section{Real-World Tasks}
\label{sec:supp_real_tasks}
\begin{figure*}[tp]
    \centering
    \includegraphics[width=\textwidth]{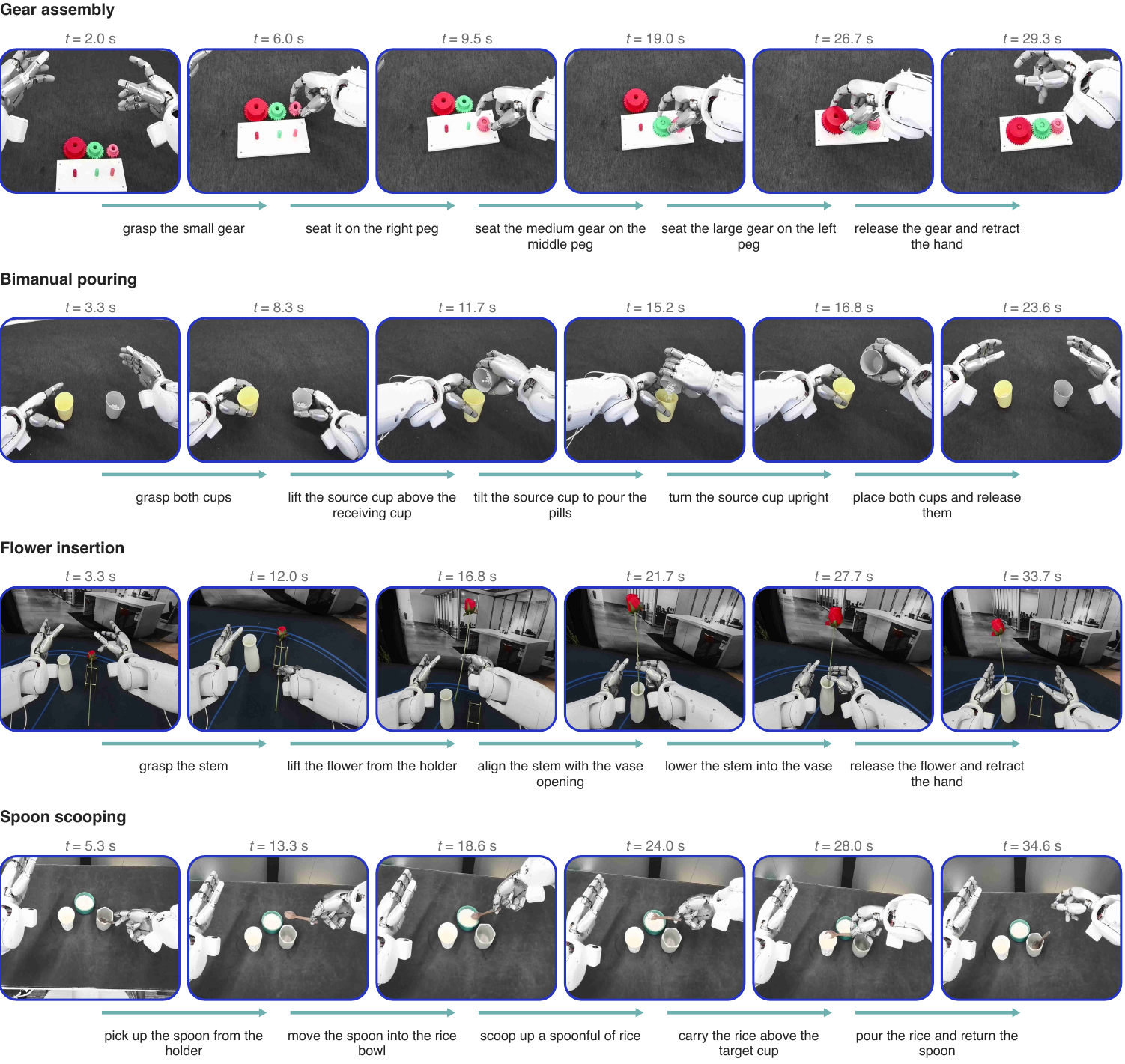}
    \caption{\textbf{Real-world tasks.} Frames of one teleoperated demonstration per task, captured by the head camera. The description between two frames is the action performed between them.}
    \label{fig:supp_real_tasks}
\end{figure*}

For each task, we describe the goal and the success criterion.
\Cref{fig:supp_real_tasks} shows one teleoperated demonstration of each task.

\paragraph{Gear assembly.}
The robot picks up three loose gears and assembles them onto the corresponding pegs of a base plate.
A trial succeeds if all three gears are fully seated on the correct pegs with adjacent teeth engaged and remain in place after the robot releases them.

\paragraph{Bimanual pouring.}
The robot transfers pills from a source cup into a receiving cup with both hands: one hand holds the receiving cup steady while the other lifts and tilts the source cup.
A trial succeeds if all pills are transferred into the receiving cup without visible spillage and both cups are placed upright on the table and released.

\paragraph{Flower insertion.}
The robot grasps a single flower by its stem, lifts it from a holder, and inserts the stem into a vase.
A trial succeeds if the stem is inside the vase and the flower remains supported by the upright vase after the robot releases it.

\paragraph{Spoon scooping.}
The robot picks up a spoon from its holder, scoops rice from a bowl, and pours a spoonful into a target cup.
A trial succeeds if a spoonful of rice is deposited into the target cup without visible spillage outside the containers and the spoon is returned to its holder and released.

%% file: main.bbl
\begin{thebibliography}{64}
\providecommand{\natexlab}[1]{#1}
\providecommand{\url}[1]{\texttt{#1}}
\expandafter\ifx\csname urlstyle\endcsname\relax
  \providecommand{\doi}[1]{doi: #1}\else
  \providecommand{\doi}{doi: \begingroup \urlstyle{rm}\Url}\fi

\bibitem[Bai et~al.(2025)Bai, Cai, Chen, Chen, Chen, Cheng, Deng, Ding, Gao,
  Ge, Ge, Guo, Huang, Huang, Huang, Hui, Jiang, Li, Li, Li, Li, Lin, Lin, Liu,
  Liu, Liu, Liu, Liu, Liu, Lu, Luo, Lv, Men, Meng, Ren, Ren, Song, Sun, Tang,
  Tu, Wan, Wang, Wang, Wang, Wang, Xie, Xu, Xu, Xu, Yang, Yang, Yang, Yang, Yu,
  Zhang, Zhang, Zhang, Zheng, Zhong, Zhou, Zhou, Zhou, Zhu, and Zhu]{Qwen3-VL}
Shuai Bai, Yuxuan Cai, Ruizhe Chen, Keqin Chen, Xionghui Chen, Zesen Cheng,
  Lianghao Deng, Wei Ding, Chang Gao, Chunjiang Ge, Wenbin Ge, Zhifang Guo,
  Qidong Huang, Jie Huang, Fei Huang, Binyuan Hui, Shutong Jiang, Zhaohai Li,
  Mingsheng Li, Mei Li, Kaixin Li, Zicheng Lin, Junyang Lin, Xuejing Liu,
  Jiawei Liu, Chenglong Liu, Yang Liu, Dayiheng Liu, Shixuan Liu, Dunjie Lu,
  Ruilin Luo, Chenxu Lv, Rui Men, Lingchen Meng, Xuancheng Ren, Xingzhang Ren,
  Sibo Song, Yuchong Sun, Jun Tang, Jianhong Tu, Jianqiang Wan, Peng Wang,
  Pengfei Wang, Qiuyue Wang, Yuxuan Wang, Tianbao Xie, Yiheng Xu, Haiyang Xu,
  Jin Xu, Zhibo Yang, Mingkun Yang, Jianxin Yang, An~Yang, Bowen Yu, Fei Zhang,
  Hang Zhang, Xi~Zhang, Bo~Zheng, Humen Zhong, Jingren Zhou, Fan Zhou, Jing
  Zhou, Yuanzhi Zhu, and Ke~Zhu.
\newblock {Qwen3-VL} technical report.
\newblock \emph{arXiv preprint arXiv:2511.21631}, 2025.

\bibitem[Baker et~al.(2022)Baker, Akkaya, Zhokhov, Huizinga, Tang, Ecoffet,
  Houghton, Sampedro, and Clune]{baker_video_2022}
Bowen Baker, Ilge Akkaya, Peter Zhokhov, Joost Huizinga, Jie Tang, Adrien
  Ecoffet, Brandon Houghton, Raul Sampedro, and Jeff Clune.
\newblock Video {PreTraining} ({VPT}): {Learning} to {Act} by {Watching}
  {Unlabeled} {Online} {Videos}, June 2022.
\newblock URL \url{http://arxiv.org/abs/2206.11795}.
\newblock arXiv:2206.11795 [cs.LG].

\bibitem[Bauer et~al.(2026)Bauer, Nava, and Katzschmann]{bauer_latent_2026}
Erik Bauer, Elvis Nava, and Robert~K. Katzschmann.
\newblock Latent {Action} {Diffusion} for {Cross}-{Embodiment} {Manipulation},
  March 2026.
\newblock URL \url{https://arxiv.org/abs/2506.14608v4}.
\newblock arXiv:2506.14608v4 [cs.RO].

\bibitem[Black et~al.(2024)Black, Brown, Driess, Esmail, Equi, Finn, Fusai,
  Groom, Hausman, Ichter, Jakubczak, Jones, Ke, Levine, Li-Bell, Mothukuri,
  Nair, Pertsch, Shi, Tanner, Vuong, Walling, Wang, and
  Zhilinsky]{black2024pi_0}
Kevin Black, Noah Brown, Danny Driess, Adnan Esmail, Michael Equi, Chelsea
  Finn, Niccolo Fusai, Lachy Groom, Karol Hausman, Brian Ichter, Szymon
  Jakubczak, Tim Jones, Liyiming Ke, Sergey Levine, Adrian Li-Bell, Mohith
  Mothukuri, Suraj Nair, Karl Pertsch, Lucy~Xiaoyang Shi, James Tanner, Quan
  Vuong, Anna Walling, Haohuan Wang, and Ury Zhilinsky.
\newblock {$\pi_0$}: A vision-language-action flow model for general robot
  control.
\newblock \emph{arXiv preprint arXiv:2410.24164}, 2024.
\newblock URL \url{https://arxiv.org/abs/2410.24164v1}.

\bibitem[Brohan et~al.(2023)Brohan, Brown, Carbajal, Chebotar, Chen,
  Choromanski, Ding, Driess, Dubey, Finn, Florence, Fu, Arenas, Gopalakrishnan,
  Han, Hausman, Herzog, Hsu, Ichter, Irpan, Joshi, Julian, Kalashnikov, Kuang,
  Leal, Lee, Lee, Levine, Lu, Michalewski, Mordatch, Pertsch, Rao, Reymann,
  Ryoo, Salazar, Sanketi, Sermanet, Singh, Singh, Soricut, Tran, Vanhoucke,
  Vuong, Wahid, Welker, Wohlhart, Wu, Xia, Xiao, Xu, Xu, Yu, and
  Zitkovich]{brohan2023rt}
Anthony Brohan, Noah Brown, Justice Carbajal, Yevgen Chebotar, Xi~Chen,
  Krzysztof Choromanski, Tianli Ding, Danny Driess, Avinava Dubey, Chelsea
  Finn, Pete Florence, Chuyuan Fu, Montse~Gonzalez Arenas, Keerthana
  Gopalakrishnan, Kehang Han, Karol Hausman, Alexander Herzog, Jasmine Hsu,
  Brian Ichter, Alex Irpan, Nikhil Joshi, Ryan Julian, Dmitry Kalashnikov,
  Yuheng Kuang, Isabel Leal, Lisa Lee, Tsang-Wei~Edward Lee, Sergey Levine, Yao
  Lu, Henryk Michalewski, Igor Mordatch, Karl Pertsch, Kanishka Rao, Krista
  Reymann, Michael Ryoo, Grecia Salazar, Pannag Sanketi, Pierre Sermanet,
  Jaspiar Singh, Anikait Singh, Radu Soricut, Huong Tran, Vincent Vanhoucke,
  Quan Vuong, Ayzaan Wahid, Stefan Welker, Paul Wohlhart, Jialin Wu, Fei Xia,
  Ted Xiao, Peng Xu, Sichun Xu, Tianhe Yu, and Brianna Zitkovich.
\newblock {RT-2}: Vision-language-action models transfer web knowledge to
  robotic control, 2023.
\newblock URL \url{https://arxiv.org/abs/2307.15818}.

\bibitem[Bruce et~al.(2024)Bruce, Dennis, Edwards, Parker-Holder, Shi, Hughes,
  Lai, Mavalankar, Steigerwald, Apps, Aytar, Bechtle, Behbahani, Chan, Heess,
  Gonzalez, Osindero, Ozair, Reed, Zhang, Zolna, Clune, de~Freitas, Singh, and
  Rockt{\"a}schel]{bruce2024genie}
Jake Bruce, Michael Dennis, Ashley Edwards, Jack Parker-Holder, Yuge Shi,
  Edward Hughes, Matthew Lai, Aditi Mavalankar, Richie Steigerwald, Chris Apps,
  Yusuf Aytar, Sarah Bechtle, Feryal Behbahani, Stephanie Chan, Nicolas Heess,
  Lucy Gonzalez, Simon Osindero, Sherjil Ozair, Scott Reed, Jingwei Zhang,
  Konrad Zolna, Jeff Clune, Nando de~Freitas, Satinder Singh, and Tim
  Rockt{\"a}schel.
\newblock {Genie}: Generative interactive environments.
\newblock \emph{arXiv preprint arXiv:2402.15391}, 2024.
\newblock URL \url{https://arxiv.org/abs/2402.15391}.

\bibitem[Bu et~al.(2025)Bu, Yang, Cai, Gao, Ren, Yao, Luo, and
  Li]{bu2025univla}
Qingwen Bu, Yanting Yang, Jisong Cai, Shenyuan Gao, Guanghui Ren, Maoqing Yao,
  Ping Luo, and Hongyang Li.
\newblock {UniVLA}: Learning to act anywhere with task-centric latent actions.
\newblock \emph{arXiv preprint arXiv:2505.06111}, 2025.

\bibitem[{Bytedance Seed}(2026)]{bytedance2026seed2}
{Bytedance Seed}.
\newblock {Seed2.0} model card: Towards intelligence frontier for real-world
  complexity, 2026.
\newblock URL \url{https://arxiv.org/abs/2607.00248}.

\bibitem[Chen et~al.(2026{\natexlab{a}})Chen, Chen, Qiu, Bai, Ge, and
  Ge]{chen2026unit}
Boyu Chen, Yi~Chen, Lu~Qiu, Jerry Bai, Yuying Ge, and Yixiao Ge.
\newblock {UniT: Toward a Unified Physical Language for Human-to-Humanoid
  Policy Learning and World Modeling}, 2026{\natexlab{a}}.
\newblock URL \url{https://arxiv.org/abs/2604.19734}.

\bibitem[Chen et~al.(2025{\natexlab{a}})Chen, Wang, Li, Sun, Chen, Liu, Wang,
  Raj, Liu, and Barsoum]{Chen_2025_CVPR}
Hao Chen, Ze~Wang, Xiang Li, Ximeng Sun, Fangyi Chen, Jiang Liu, Jindong Wang,
  Bhiksha Raj, Zicheng Liu, and Emad Barsoum.
\newblock {SoftVQ-VAE}: Efficient 1-dimensional continuous tokenizer.
\newblock In \emph{Proceedings of the IEEE/CVF Conference on Computer Vision
  and Pattern Recognition (CVPR)}, pages 28358--28370, June 2025{\natexlab{a}}.

\bibitem[Chen et~al.(2025{\natexlab{b}})Chen, Chen, Chen, Cai, Liu, Li, Liang,
  Lin, Ge, Gu, Deng, Guo, Nian, Xie, Chen, Su, Xu, Liu, Hu, Gao, Wang, Liang,
  Qin, Yang, Luo, and Mu]{chen2025robotwin}
Tianxing Chen, Zanxin Chen, Baijun Chen, Zijian Cai, Yibin Liu, Zixuan Li,
  Qiwei Liang, Xianliang Lin, Yiheng Ge, Zhenyu Gu, Weiliang Deng, Yubin Guo,
  Tian Nian, Xuanbing Xie, Qiangyu Chen, Kailun Su, Tianling Xu, Guodong Liu,
  Mengkang Hu, Huan-ang Gao, Kaixuan Wang, Zhixuan Liang, Yusen Qin, Xiaokang
  Yang, Ping Luo, and Yao Mu.
\newblock {RoboTwin} 2.0: A scalable data generator and benchmark with strong
  domain randomization for robust bimanual robotic manipulation.
\newblock \emph{arXiv preprint arXiv:2506.18088}, 2025{\natexlab{b}}.

\bibitem[Chen et~al.(2024{\natexlab{a}})Chen, Guo, He, Zhang, Zhang, Yang,
  Zhao, and Bian]{chen_igor_2024}
Xiaoyu Chen, Junliang Guo, Tianyu He, Chuheng Zhang, Pushi Zhang, Derek~Cathera
  Yang, Li~Zhao, and Jiang Bian.
\newblock {IGOR}: {Image}-{GOal} {Representations} are the {Atomic} {Control}
  {Units} for {Foundation} {Models} in {Embodied} {AI}, October
  2024{\natexlab{a}}.
\newblock URL \url{http://arxiv.org/abs/2411.00785}.
\newblock arXiv:2411.00785 [cs.RO].

\bibitem[Chen et~al.(2026{\natexlab{b}})Chen, Wei, Zhang, Zhang, Wang, Guo,
  Yang, Wang, Xiao, Zhao, Chen, and Bian]{chen2026iclr-villax}
Xiaoyu Chen, Hangxing Wei, Pushi Zhang, Chuheng Zhang, Kaixin Wang, Yanjiang
  Guo, Rushuai Yang, Yucen Wang, Xinquan Xiao, Li~Zhao, Jianyu Chen, and Jiang
  Bian.
\newblock {villa-X}: Enhancing latent action modeling in vision-language-action
  models.
\newblock In \emph{International Conference on Learning Representations},
  2026{\natexlab{b}}.
\newblock URL
  \url{https://proceedings.iclr.cc/paper_files/paper/2026/hash/73471f46f69a13cede6cf2db3a6f438f-Abstract-Conference.html}.

\bibitem[Chen et~al.(2024{\natexlab{b}})Chen, Ge, Tang, Li, Ge, Ding, Shan, and
  Liu]{chen2024moto}
Yi~Chen, Yuying Ge, Weiliang Tang, Yizhuo Li, Yixiao Ge, Mingyu Ding, Ying
  Shan, and Xihui Liu.
\newblock {Moto}: Latent motion token as the bridging language for learning
  robot manipulation from videos.
\newblock \emph{arXiv preprint arXiv:2412.04445}, 2024{\natexlab{b}}.

\bibitem[Dai et~al.(2026)Dai, Lan, Zhou, Zhao, Su, Tong, Guan, and
  Yang]{dai_conla_2026}
Weisheng Dai, Kai Lan, Jianyi Zhou, Bo~Zhao, Xiu Su, Junwen Tong, Weili Guan,
  and Shuo Yang.
\newblock {ConLA}: {Contrastive} {Latent} {Action} {Learning} from {Human}
  {Videos} for {Robotic} {Manipulation}, January 2026.
\newblock URL \url{http://arxiv.org/abs/2602.00557}.
\newblock arXiv:2602.00557 [cs.RO].

\bibitem[Du et~al.(2023)Du, Yang, Dai, Dai, Nachum, Tenenbaum, Schuurmans, and
  Abbeel]{du2023unipi}
Yilun Du, Mengjiao Yang, Bo~Dai, Hanjun Dai, Ofir Nachum, Joshua~B. Tenenbaum,
  Dale Schuurmans, and Pieter Abbeel.
\newblock {Learning Universal Policies via Text-Guided Video Generation}, 2023.
\newblock URL \url{https://arxiv.org/abs/2302.00111}.

\bibitem[Fan et~al.(2026)Fan, Wu, Che, Wang, Wu, Liao, Liu, Zhang, Zhao, Xu,
  Li, Liu, Zhang, Wan, and Tang]{fan_xr-1_2026}
Shichao Fan, Kun Wu, Zhengping Che, Xinhua Wang, Di~Wu, Fei Liao, Ning Liu,
  Yixue Zhang, Zhen Zhao, Zhiyuan Xu, Meng Li, Qingjie Liu, Shanghang Zhang,
  Min Wan, and Jian Tang.
\newblock {XR}-1: {Towards} {Versatile} {Vision}-{Language}-{Action} {Models}
  via {Learning} {Unified} {Vision}-{Motion} {Representations}, July 2026.
\newblock URL \url{https://arxiv.org/abs/2511.02776v3}.
\newblock arXiv:2511.02776v3 [cs.RO].

\bibitem[Fu et~al.(2025)Fu, Chen, Zhao, Shan, Yao, Wang, Wang, and
  Zhang]{fu_metis_2025}
Yankai Fu, Ning Chen, Junkai Zhao, Shaozhe Shan, Guocai Yao, Pengwei Wang,
  Zhongyuan Wang, and Shanghang Zhang.
\newblock {METIS}: {Multi}-{Source} {Egocentric} {Training} for {Integrated}
  {Dexterous} {Vision}-{Language}-{Action} {Model}, November 2025.
\newblock URL \url{http://arxiv.org/abs/2511.17366}.
\newblock arXiv:2511.17366 [cs.RO].

\bibitem[Grauman et~al.(2022)Grauman, Westbury, Byrne, Chavis, Furnari,
  Girdhar, Hamburger, Jiang, Liu, Liu, Martin, Nagarajan, Radosavovic,
  Ramakrishnan, Ryan, Sharma, Wray, Xu, Xu, Zhao, Bansal, Batra, Cartillier,
  Crane, Do, Doulaty, Erapalli, Feichtenhofer, Fragomeni, Fu, Gebreselasie,
  Gonz\'alez, Hillis, Huang, Huang, Jia, Khoo, Kol\'a\v{r}, Kottur, Kumar,
  Landini, Li, Li, Li, Mangalam, Modhugu, Munro, Murrell, Nishiyasu, Price,
  Ruiz, Ramazanova, Sari, Somasundaram, Southerland, Sugano, Tao, Vo, Wang, Wu,
  Yagi, Zhao, Zhu, Arbel\'aez, Crandall, Damen, Farinella, Fuegen, Ghanem,
  Ithapu, Jawahar, Joo, Kitani, Li, Newcombe, Oliva, Park, Rehg, Sato, Shi,
  Shou, Torralba, Torresani, Yan, and Malik]{Grauman_2022_CVPR}
Kristen Grauman, Andrew Westbury, Eugene Byrne, Zachary Chavis, Antonino
  Furnari, Rohit Girdhar, Jackson Hamburger, Hao Jiang, Miao Liu, Xingyu Liu,
  Miguel Martin, Tushar Nagarajan, Ilija Radosavovic, Santhosh~Kumar
  Ramakrishnan, Fiona Ryan, Jayant Sharma, Michael Wray, Mengmeng Xu,
  Eric~Zhongcong Xu, Chen Zhao, Siddhant Bansal, Dhruv Batra, Vincent
  Cartillier, Sean Crane, Tien Do, Morrie Doulaty, Akshay Erapalli, Christoph
  Feichtenhofer, Adriano Fragomeni, Qichen Fu, Abrham Gebreselasie, Cristina
  Gonz\'alez, James Hillis, Xuhua Huang, Yifei Huang, Wenqi Jia, Weslie Khoo,
  J\'achym Kol\'a\v{r}, Satwik Kottur, Anurag Kumar, Federico Landini, Chao Li,
  Yanghao Li, Zhenqiang Li, Karttikeya Mangalam, Raghava Modhugu, Jonathan
  Munro, Tullie Murrell, Takumi Nishiyasu, Will Price, Paola Ruiz, Merey
  Ramazanova, Leda Sari, Kiran Somasundaram, Audrey Southerland, Yusuke Sugano,
  Ruijie Tao, Minh Vo, Yuchen Wang, Xindi Wu, Takuma Yagi, Ziwei Zhao, Yunyi
  Zhu, Pablo Arbel\'aez, David Crandall, Dima Damen, Giovanni~Maria Farinella,
  Christian Fuegen, Bernard Ghanem, Vamsi~Krishna Ithapu, C.~V. Jawahar,
  Hanbyul Joo, Kris Kitani, Haizhou Li, Richard Newcombe, Aude Oliva, Hyun~Soo
  Park, James~M. Rehg, Yoichi Sato, Jianbo Shi, Mike~Zheng Shou, Antonio
  Torralba, Lorenzo Torresani, Mingfei Yan, and Jitendra Malik.
\newblock {Ego4D}: Around the world in 3,000 hours of egocentric video.
\newblock In \emph{Proceedings of the IEEE/CVF Conference on Computer Vision
  and Pattern Recognition (CVPR)}, pages 18995--19012, June 2022.
\newblock URL
  \url{https://openaccess.thecvf.com/content/CVPR2022/html/Grauman_Ego4D_Around_the_World_in_3000_Hours_of_Egocentric_Video_CVPR_2022_paper.html}.

\bibitem[Hoque et~al.(2026)Hoque, Huang, Yoon, Sivapurapu, and
  Zhang]{hoque2026egodex}
Ryan Hoque, Peide Huang, David~J. Yoon, Mouli Sivapurapu, and Jian Zhang.
\newblock {EgoDex}: Learning dexterous manipulation from large-scale egocentric
  video.
\newblock In \emph{International Conference on Learning Representations}, 2026.
\newblock URL \url{https://openreview.net/forum?id=FFxkFMU89E}.

\bibitem[Kareer et~al.(2024)Kareer, Patel, Punamiya, Mathur, Cheng, Wang,
  Hoffman, and Xu]{kareer2024egomimic}
Simar Kareer, Dhruv Patel, Ryan Punamiya, Pranay Mathur, Shuo Cheng, Chen Wang,
  Judy Hoffman, and Danfei Xu.
\newblock {EgoMimic: Scaling Imitation Learning via Egocentric Video}, 2024.
\newblock URL \url{https://arxiv.org/abs/2410.24221}.

\bibitem[Kareer et~al.(2025)Kareer, Pertsch, Darpinian, Hoffman, Xu, Levine,
  Finn, and Nair]{kareer_emergence_2025}
Simar Kareer, Karl Pertsch, James Darpinian, Judy Hoffman, Danfei Xu, Sergey
  Levine, Chelsea Finn, and Suraj Nair.
\newblock Emergence of {Human} to {Robot} {Transfer} in
  {Vision}-{Language}-{Action} {Models}, December 2025.
\newblock URL \url{http://arxiv.org/abs/2512.22414}.
\newblock arXiv:2512.22414 [cs.RO].

\bibitem[Kim et~al.(2024)Kim, Pertsch, Karamcheti, Xiao, Balakrishna, Nair,
  Rafailov, Foster, Lam, Sanketi, Vuong, Kollar, Burchfiel, Tedrake, Sadigh,
  Levine, Liang, and Finn]{kim2024openvla}
Moo~Jin Kim, Karl Pertsch, Siddharth Karamcheti, Ted Xiao, Ashwin Balakrishna,
  Suraj Nair, Rafael Rafailov, Ethan Foster, Grace Lam, Pannag Sanketi, Quan
  Vuong, Thomas Kollar, Benjamin Burchfiel, Russ Tedrake, Dorsa Sadigh, Sergey
  Levine, Percy Liang, and Chelsea Finn.
\newblock {OpenVLA}: An open-source vision-language-action model, 2024.
\newblock URL \url{https://arxiv.org/abs/2406.09246}.

\bibitem[Kingma and Welling(2013)]{kingma2013auto}
Diederik~P Kingma and Max Welling.
\newblock Auto-encoding variational {Bayes}.
\newblock \emph{arXiv preprint arXiv:1312.6114}, 2013.
\newblock URL \url{https://arxiv.org/abs/1312.6114}.

\bibitem[Ko et~al.(2023)Ko, Mao, Du, Sun, and Tenenbaum]{ko2023avdc}
Po-Chen Ko, Jiayuan Mao, Yilun Du, Shao-Hua Sun, and Joshua~B. Tenenbaum.
\newblock {Learning to Act from Actionless Videos through Dense
  Correspondences}, 2023.
\newblock URL \url{https://arxiv.org/abs/2310.08576}.

\bibitem[Li et~al.(2026{\natexlab{a}})Li, Yin, Lin, Zhang, and
  Xu]{li_egowam_2026}
Baoyu Li, Xinchen Yin, Mengying Lin, Yixin Zhang, and Danfei Xu.
\newblock {EgoWAM}: {World} {Action} {Models} {Beyond} {Pixels} with
  {In}-the-{Wild} {Egocentric} {Human} {Data}, July 2026{\natexlab{a}}.
\newblock URL \url{http://arxiv.org/abs/2607.08436}.
\newblock arXiv:2607.08436 [cs.RO].

\bibitem[Li et~al.(2026{\natexlab{b}})Li, Zhang, Luo, Yang, Wang, Han, Yu, Gao,
  Xue, Zhu, Shen, and Xu]{li2026lingbotva}
Lin Li, Qihang Zhang, Yiming Luo, Shuai Yang, Ruilin Wang, Fei Han, Mingrui Yu,
  Zelin Gao, Nan Xue, Xing Zhu, Yujun Shen, and Yinghao Xu.
\newblock {Causal World Modeling for Robot Control}, 2026{\natexlab{b}}.
\newblock URL \url{https://arxiv.org/abs/2601.21998}.

\bibitem[Li et~al.(2026{\natexlab{c}})Li, Gong, Li, Li, Zhou, Ye, Zhou, and
  Mu]{li2026rotvla}
Qiwei Li, Xicheng Gong, Xinghang Li, Peiyan Li, Quanyun Zhou, Hangjun Ye,
  Jiahuan Zhou, and Yadong Mu.
\newblock {RotVLA: Rotational Latent Action for Vision-Language-Action Model},
  2026{\natexlab{c}}.
\newblock URL \url{https://arxiv.org/abs/2605.13403}.

\bibitem[Liang et~al.(2026)Liang, Czempin, Hong, Zhou, Wang, Biyik, and
  Tu]{liang_clam_2026}
Anthony Liang, Pavel Czempin, Matthew~M. Hong, Yutai Zhou, Jingzhen Wang, Erdem
  Biyik, and Stephen Tu.
\newblock {CLAM}: {Continuous} {Latent} {Action} {Models} for {Robot}
  {Learning} from {Unlabeled} {Demonstrations}, July 2026.
\newblock URL \url{https://arxiv.org/abs/2505.04999v2}.
\newblock arXiv:2505.04999v2 [cs.RO].

\bibitem[Lin et~al.(2026)Lin, Du, Mao, Ye, Zhong, Cheng, Wang, Liu, Tian, Yan,
  Wu, Meng, Wei, Fu, Li, and Zhao]{lin_la4vla_2026}
Tao Lin, Yuxin Du, Yiran Mao, Zewei Ye, Yilei Zhong, Bing Cheng, Yiming Wang,
  Jiting Liu, Yang Tian, Junchi Yan, Feiran Wu, Zenan Meng, Hu~Wei, Yuqian Fu,
  Gen Li, and Bo~Zhao.
\newblock {LA4VLA}: {Learning} to {Act} without {Seeing} via
  {Language}-{Action} {Pretraining}, June 2026.
\newblock URL \url{http://arxiv.org/abs/2606.27295}.
\newblock arXiv:2606.27295 [cs.RO].

\bibitem[Liu et~al.(2026)Liu, Feng, Fu, Wang, Lu, and Huang]{liu_scar_2026}
Hongjia Liu, Fan Feng, Minghao Fu, Xinyue Wang, Haofei Lu, and Biwei Huang.
\newblock {SCAR}: {Self}-{Supervised} {Continuous} {Action} {Representation}
  {Learning}, May 2026.
\newblock URL \url{http://arxiv.org/abs/2605.16412}.
\newblock arXiv:2605.16412 [cs.RO].

\bibitem[Luo et~al.(2025)Luo, Feng, Zhang, Zheng, Wang, Yuan, Liu, Xu, Jin, and
  Lu]{luo2025beingh0}
Hao Luo, Yicheng Feng, Wanpeng Zhang, Sipeng Zheng, Ye~Wang, Haoqi Yuan,
  Jiazheng Liu, Chaoyi Xu, Qin Jin, and Zongqing Lu.
\newblock {Being-H0: Vision-Language-Action Pretraining from Large-Scale Human
  Videos}, 2025.
\newblock URL \url{https://arxiv.org/abs/2507.15597}.

\bibitem[Luo et~al.(2026)Luo, Wang, Zhang, Yuan, Feng, Xu, Zheng, and
  Lu]{luo_joint-aligned_2026}
Hao Luo, Ye~Wang, Wanpeng Zhang, Haoqi Yuan, Yicheng Feng, Haiweng Xu, Sipeng
  Zheng, and Zongqing Lu.
\newblock Joint-{Aligned} {Latent} {Action}: {Towards} {Scalable} {VLA}
  {Pretraining} in the {Wild}, February 2026.
\newblock URL \url{http://arxiv.org/abs/2602.21736}.
\newblock arXiv:2602.21736 [cs.RO].

\bibitem[Ma et~al.(2026)Ma, Cai, Xu, Li, Yang, Tian, Cao, Zhu, Qiu,
  Zhaxizhuoma, Yang, Peng, Wei, Zhu, Jiang, Gao, Wang, Yuan, Li, Zhu, Wang,
  Ding, Pang, Zeng, Zhang, Zhou, Mu, Shen, and Zhang]{ma2026internvlaa15}
Haoxiang Ma, Junhao Cai, Xiaoxu Xu, Hao Li, Yuyin Yang, Yang Tian, Jiafei Cao,
  Hongrui Zhu, Zherui Qiu, Zhaxizhuoma, Yuqiang Yang, Jiaqi Peng, Xueyuan Wei,
  Yangkun Zhu, Jiahao Jiang, Xing Gao, Hanqing Wang, Feng Yuan, Kailin Li,
  Xueyue Zhu, Tai Wang, Yan Ding, Jiangmiao Pang, Jia Zeng, Jingjing Zhang,
  Bowen Zhou, Yao Mu, Chunhua Shen, and Weinan Zhang.
\newblock {InternVLA-A1.5: Unifying Understanding, Latent Foresight, and Action
  for Compositional Generalization}, 2026.
\newblock URL \url{https://arxiv.org/abs/2607.04988}.

\bibitem[Miao et~al.(2026)Miao, Feng, Wu, Lin, He, Li, and
  Long]{miao_jepa-vla_2026}
Shangchen Miao, Ningya Feng, Jialong Wu, Ye~Lin, Xu~He, Dong Li, and Mingsheng
  Long.
\newblock {JEPA}-{VLA}: {Video} {Predictive} {Embedding} is {Needed} for {VLA}
  {Models}, February 2026.
\newblock URL \url{http://arxiv.org/abs/2602.11832}.
\newblock arXiv:2602.11832 [cs].

\bibitem[Nair et~al.(2022)Nair, Rajeswaran, Kumar, Finn, and
  Gupta]{nair2022r3m}
Suraj Nair, Aravind Rajeswaran, Vikash Kumar, Chelsea Finn, and Abhinav Gupta.
\newblock {R3M: A Universal Visual Representation for Robot Manipulation},
  2022.
\newblock URL \url{https://arxiv.org/abs/2203.12601}.

\bibitem[NVIDIA et~al.(2025)NVIDIA, Bjorck, Castañeda, Cherniadev, Da, Ding,
  Fan, Fang, Fox, Hu, Huang, Jang, Jiang, Kautz, Kundalia, Lao, Li, Lin, Lin,
  Liu, Llontop, Magne, Mandlekar, Narayan, Nasiriany, Reed, Tan, Wang, Wang,
  Wang, Wang, Xiang, Xie, Xu, Xu, Ye, Yu, Zhang, Zhang, Zhao, Zheng, and
  Zhu]{bjorck2025gr00t}
NVIDIA, Johan Bjorck, Fernando Castañeda, Nikita Cherniadev, Xingye Da, Runyu
  Ding, Linxi~"Jim" Fan, Yu~Fang, Dieter Fox, Fengyuan Hu, Spencer Huang, Joel
  Jang, Zhenyu Jiang, Jan Kautz, Kaushil Kundalia, Lawrence Lao, Zhiqi Li,
  Zongyu Lin, Kevin Lin, Guilin Liu, Edith Llontop, Loic Magne, Ajay Mandlekar,
  Avnish Narayan, Soroush Nasiriany, Scott Reed, You~Liang Tan, Guanzhi Wang,
  Zu~Wang, Jing Wang, Qi~Wang, Jiannan Xiang, Yuqi Xie, Yinzhen Xu, Zhenjia Xu,
  Seonghyeon Ye, Zhiding Yu, Ao~Zhang, Hao Zhang, Yizhou Zhao, Ruijie Zheng,
  and Yuke Zhu.
\newblock {GR00T N1: An Open Foundation Model for Generalist Humanoid Robots},
  2025.
\newblock URL \url{https://arxiv.org/abs/2503.14734}.

\bibitem[NVIDIA et~al.(2026)]{nvidia_cosmos_2026}
NVIDIA et~al.
\newblock Cosmos 3: {Omnimodal} {World} {Models} for {Physical} {AI}, June
  2026.
\newblock URL \url{http://arxiv.org/abs/2606.02800}.
\newblock arXiv:2606.02800 [cs.CV].

\bibitem[Oquab et~al.(2024)Oquab, Darcet, Moutakanni, Vo, Szafraniec, Khalidov,
  Fernandez, Haziza, Massa, El{-}Nouby, Assran, Ballas, Galuba, Howes, Huang,
  Li, Misra, Rabbat, Sharma, Synnaeve, Xu, J{\'{e}}gou, Mairal, Labatut,
  Joulin, and Bojanowski]{oquab2024dinov2}
Maxime Oquab, Timoth{\'{e}}e Darcet, Th{\'{e}}o Moutakanni, Huy~V. Vo, Marc
  Szafraniec, Vasil Khalidov, Pierre Fernandez, Daniel Haziza, Francisco Massa,
  Alaaeldin El{-}Nouby, Mido Assran, Nicolas Ballas, Wojciech Galuba, Russell
  Howes, Po{-}Yao Huang, Shang{-}Wen Li, Ishan Misra, Michael Rabbat, Vasu
  Sharma, Gabriel Synnaeve, Hu~Xu, Herv{\'{e}} J{\'{e}}gou, Julien Mairal,
  Patrick Labatut, Armand Joulin, and Piotr Bojanowski.
\newblock {DINOv2}: Learning robust visual features without supervision.
\newblock \emph{Transactions on Machine Learning Research}, 2024, 2024.
\newblock URL \url{https://openreview.net/forum?id=a68SUt6zFt}.

\bibitem[{Physical Intelligence} et~al.(2025){Physical Intelligence}, Black,
  Brown, Darpinian, Dhabalia, Driess, Esmail, Equi, Finn, Fusai, Galliker,
  Ghosh, Groom, Hausman, Ichter, Jakubczak, Jones, Ke, LeBlanc, Levine,
  Li-Bell, Mothukuri, Nair, Pertsch, Ren, Shi, Smith, Springenberg, Stachowicz,
  Tanner, Vuong, Walke, Walling, Wang, Yu, and Zhilinsky]{intelligence2025pi05}
{Physical Intelligence}, Kevin Black, Noah Brown, James Darpinian, Karan
  Dhabalia, Danny Driess, Adnan Esmail, Michael Equi, Chelsea Finn, Niccolo
  Fusai, Manuel~Y. Galliker, Dibya Ghosh, Lachy Groom, Karol Hausman, Brian
  Ichter, Szymon Jakubczak, Tim Jones, Liyiming Ke, Devin LeBlanc, Sergey
  Levine, Adrian Li-Bell, Mohith Mothukuri, Suraj Nair, Karl Pertsch, Allen~Z.
  Ren, Lucy~Xiaoyang Shi, Laura Smith, Jost~Tobias Springenberg, Kyle
  Stachowicz, James Tanner, Quan Vuong, Homer Walke, Anna Walling, Haohuan
  Wang, Lili Yu, and Ury Zhilinsky.
\newblock {$\pi_{0.5}$: a Vision-Language-Action Model with Open-World
  Generalization}, 2025.
\newblock URL \url{https://arxiv.org/abs/2504.16054}.

\bibitem[Punamiya et~al.(2026)Punamiya, Kareer, Liu, Citron, Qiu, Cai,
  Gavryushin, Chen, Liconti, Zhu, Aphiwetsa, Li, Cheluva, Kuppili, Liu, Patel,
  Gao, Chung, Co, Zbizika, Liu, Xu, Xiong, Chen, Oliani, Xuan, Yang, Wang,
  Fort, Newcombe, Gao, Chong, Matsuda, Doriwala, Pollefeys, Katzschmann, Wang,
  Song, Hoffman, and Xu]{punamiya2026egoverse}
Ryan Punamiya, Simar Kareer, Zeyi Liu, Josh Citron, Ri-Zhao Qiu, Xiongyi Cai,
  Alexey Gavryushin, Jiaqi Chen, Davide Liconti, Lawrence~Y. Zhu, Patcharapong
  Aphiwetsa, Baoyu Li, Aniketh Cheluva, Pranav Kuppili, Yangcen Liu, Dhruv
  Patel, Aidan Gao, Hye-Young Chung, Ryan Co, Renee Zbizika, Jeff Liu, Xiaomeng
  Xu, Haoyu Xiong, Geng Chen, Sebastiano Oliani, Wenkai Xuan, Chenyu Yang,
  Xi~Wang, James Fort, Richard Newcombe, Josh Gao, Jason Chong, Garrett
  Matsuda, Aseem Doriwala, Marc Pollefeys, Robert Katzschmann, Xiaolong Wang,
  Shuran Song, Judy Hoffman, and Danfei Xu.
\newblock {EgoVerse}: An egocentric human dataset for robot learning from
  around the world.
\newblock \emph{CoRR}, abs/2604.07607, 2026.
\newblock \doi{10.48550/ARXIV.2604.07607}.
\newblock URL \url{https://arxiv.org/abs/2604.07607v2}.

\bibitem[Qiu et~al.(2025)Qiu, Yang, Cheng, Chawla, Li, He, Yan, Yoon, Hoque,
  Paulsen, Yang, Zhang, Yi, Shi, and Wang]{qiu2025humanoid}
Ri-Zhao Qiu, Shiqi Yang, Xuxin Cheng, Chaitanya Chawla, Jialong Li, Tairan He,
  Ge~Yan, David~J. Yoon, Ryan Hoque, Lars Paulsen, Ge~Yang, Jian Zhang, Sha Yi,
  Guanya Shi, and Xiaolong Wang.
\newblock {Humanoid Policy $\sim$ Human Policy}, 2025.
\newblock URL \url{https://arxiv.org/abs/2503.13441}.

\bibitem[Radford et~al.(2021)Radford, Kim, Hallacy, Ramesh, Goh, Agarwal,
  Sastry, Askell, Mishkin, Clark, Krueger, and Sutskever]{radford2021clip}
Alec Radford, Jong~Wook Kim, Chris Hallacy, Aditya Ramesh, Gabriel Goh,
  Sandhini Agarwal, Girish Sastry, Amanda Askell, Pamela Mishkin, Jack Clark,
  Gretchen Krueger, and Ilya Sutskever.
\newblock Learning transferable visual models from natural language
  supervision.
\newblock In \emph{Proceedings of the 38th International Conference on Machine
  Learning}, volume 139 of \emph{Proceedings of Machine Learning Research},
  pages 8748--8763. PMLR, 2021.
\newblock URL \url{http://proceedings.mlr.press/v139/radford21a.html}.

\bibitem[Schmidt and Jiang(2023)]{schmidt2023learning}
Dominik Schmidt and Minqi Jiang.
\newblock Learning to act without actions.
\newblock \emph{arXiv preprint arXiv:2312.10812}, 2023.

\bibitem[Su et~al.(2026)Su, Chen, Shi, Liu, Zhang, Huang, Zhong, Zhu, Liu, and
  Liu]{su_world_2026}
Yue Su, Sijin Chen, Haixin Shi, Mingyu Liu, Zhengshen Zhang, Ningyuan Huang,
  Weiheng Zhong, Zhengbang Zhu, Yuxiao Liu, and Xihui Liu.
\newblock World {Guidance}: {World} {Modeling} in {Condition} {Space} for
  {Action} {Generation}, February 2026.
\newblock URL \url{http://arxiv.org/abs/2602.22010}.
\newblock arXiv:2602.22010 [cs.RO].

\bibitem[Tang et~al.(2026)Tang, Xie, Huang, Sun, Yang, Luo, Jin, Fang, and
  Wang]{tang_towards_2026}
Peijun Tang, Shangjin Xie, Baifu Huang, Binyan Sun, Haotian Yang, Kuncheng Luo,
  Weiqi Jin, Shilin Fang, and Jianan Wang.
\newblock Towards {Predictive}, {Aligned}, and {Scalable} {Robot} {Learning},
  July 2026.
\newblock URL \url{http://arxiv.org/abs/2607.11270}.
\newblock arXiv:2607.11270 [cs.RO].

\bibitem[van~den Oord et~al.(2017)van~den Oord, Vinyals, and
  Kavukcuoglu]{vandenoord2017neurips-neural}
Aaron van~den Oord, Oriol Vinyals, and Koray Kavukcuoglu.
\newblock Neural discrete representation learning.
\newblock In \emph{Advances in Neural Information Processing Systems},
  volume~30, pages 6306--6315, 2017.
\newblock URL
  \url{https://proceedings.neurips.cc/paper_files/paper/2017/hash/7a98af17e63a0ac09ce2e96d03992fbc-Abstract.html}.

\bibitem[Wang et~al.(2023)Wang, Fan, Sun, Zhang, Fei-Fei, Xu, Zhu, and
  Anandkumar]{wang2023mimicplay}
Chen Wang, Linxi Fan, Jiankai Sun, Ruohan Zhang, Li~Fei-Fei, Danfei Xu, Yuke
  Zhu, and Anima Anandkumar.
\newblock {MimicPlay: Long-Horizon Imitation Learning by Watching Human Play},
  2023.
\newblock URL \url{https://arxiv.org/abs/2302.12422}.

\bibitem[Wang et~al.(2026)Wang, Huang, Li, Zhang, Liang, Jin, Chen, Chi, Zhou,
  Yu, Wang, Rui, Yao, Yuan, Shen, Zhu, Zhu, Gao, Chi, He, Zhang, Dong, Shao,
  and Zhao]{wang2026openwam}
Yuran Wang, Siqiao Huang, Mingleyang Li, Chenhao Zhang, Jiaqi Liang, Weiyang
  Jin, Yue Chen, Xuemin Chi, Donghao Zhou, Qize Yu, Yu-Kai Wang, Yuhan Rui,
  Shenzhe Yao, Zhen Yuan, Zhenhao Shen, Kefei Zhu, Zijie Zhu, Ning Gao, Xiaowei
  Chi, Guanqi He, Shanghang Zhang, Hao Dong, Lin Shao, and Hang Zhao.
\newblock {OpenWAM}: An open, modular exploration towards systematic
  world-action model pretraining.
\newblock \emph{arXiv preprint arXiv:2609.07398}, 2026.

\bibitem[Weng et~al.(2026)Weng, Wu, Chen, Xie, Wang, Xu, Song, and
  Shen]{weng_language-grounded_2026}
Wuding Weng, Tongshu Wu, Liucheng Chen, Siyu Xie, Zheng Wang, Xing Xu, Jingkuan
  Song, and Heng~Tao Shen.
\newblock Language-{Grounded} {Decoupled} {Action} {Representation} for
  {Robotic} {Manipulation}, March 2026.
\newblock URL \url{http://arxiv.org/abs/2603.12967}.
\newblock arXiv:2603.12967 [cs.RO].

\bibitem[Wu et~al.(2023)Wu, Jing, Cheang, Chen, Xu, Li, Liu, Li, and
  Kong]{wu2023gr1}
Hongtao Wu, Ya~Jing, Chilam Cheang, Guangzeng Chen, Jiafeng Xu, Xinghang Li,
  Minghuan Liu, Hang Li, and Tao Kong.
\newblock {Unleashing Large-Scale Video Generative Pre-training for Visual
  Robot Manipulation}, 2023.
\newblock URL \url{https://arxiv.org/abs/2312.13139}.

\bibitem[Yang et~al.(2025)Yang, Yu, Wu, Yan, Li, Cheng, Zou, Fang, Cheng, Qiu,
  Yin, Liu, Han, Lu, and Wang]{yang2025egovla}
Ruihan Yang, Qinxi Yu, Yecheng Wu, Rui Yan, Borui Li, An-Chieh Cheng, Xueyan
  Zou, Yunhao Fang, Xuxin Cheng, Ri-Zhao Qiu, Hongxu Yin, Sifei Liu, Song Han,
  Yao Lu, and Xiaolong Wang.
\newblock {EgoVLA: Learning Vision-Language-Action Models from Egocentric Human
  Videos}, 2025.
\newblock URL \url{https://arxiv.org/abs/2507.12440}.

\bibitem[Yang et~al.(2026)Yang, Liu, Kou, Chen, Hu, Zhou, Zhao, Wei, Xia, Li,
  Liu, and Deng]{yang_world-language-action_2026}
Yi~Yang, Zhihong Liu, Siqi Kou, Yiyang Chen, Yanzhe Hu, Jianbo Zhou, Boyuan
  Zhao, Zhijie Wei, Xiao Xia, Xueqi Li, Pengfei Liu, and Zhijie Deng.
\newblock World-{Language}-{Action} {Model} for {Unified} {World} {Modeling},
  {Language} {Reasoning}, and {Action} {Synthesis}, June 2026.
\newblock URL \url{http://arxiv.org/abs/2606.05979}.
\newblock arXiv:2606.05979 [cs.RO].

\bibitem[Ye et~al.(2025)Ye, Jang, Jeon, Joo, Yang, Peng, Mandlekar, Tan, Chao,
  Lin, Liden, Lee, Gao, Zettlemoyer, Fox, and Seo]{ye2025latent}
Seonghyeon Ye, Joel Jang, Byeongguk Jeon, Se~June Joo, Jianwei Yang, Baolin
  Peng, Ajay Mandlekar, Reuben Tan, Yu-Wei Chao, Bill~Yuchen Lin, Lars Liden,
  Kimin Lee, Jianfeng Gao, Luke Zettlemoyer, Dieter Fox, and Minjoon Seo.
\newblock Latent action pretraining from videos.
\newblock In \emph{International Conference on Learning Representations}, 2025.
\newblock URL
  \url{https://proceedings.iclr.cc/paper_files/paper/2025/hash/45d74e190008c7bff2845ffc8e3facd3-Abstract-Conference.html}.

\bibitem[Yuan et~al.(2026{\natexlab{a}})Yuan, Liang, Chen, Wang, Li, Lin,
  Huang, Lei, Zhang, Zhang, Zhang, Fan, Zhou, Peng, Lv, Chen, Yang, Huang, Lin,
  Liu, Zhou, Wu, and Chen]{yuan_qwen-robotmanip_2026}
Haoqi Yuan, Zhixuan Liang, Anzhe Chen, Ye~Wang, Haoyang Li, Pei Lin, Yiyang
  Huang, Zixing Lei, Tong Zhang, Jiazhao Zhang, Jie Zhang, Jingyang Fan, Gengze
  Zhou, Qihang Peng, Chenxu Lv, Xiaoyue Chen, An~Yang, Fei Huang, Junyang Lin,
  Dayiheng Liu, Jingren Zhou, Chenfei Wu, and Xiong-Hui Chen.
\newblock Qwen-{RobotManip} {Technical} {Report}: {Alignment} {Unlocks} {Scale}
  for {Robotic} {Manipulation} {Foundation} {Models}, June 2026{\natexlab{a}}.
\newblock URL \url{http://arxiv.org/abs/2606.17846}.
\newblock arXiv:2606.17846 [cs.RO].

\bibitem[Yuan et~al.(2026{\natexlab{b}})Yuan, Dong, Liu, and
  Zhao]{yuan2026fastwam}
Tianyuan Yuan, Zibin Dong, Yicheng Liu, and Hang Zhao.
\newblock {Fast-WAM: Do World Action Models Need Test-time Future
  Imagination?}, 2026{\natexlab{b}}.
\newblock URL \url{https://arxiv.org/abs/2603.16666}.

\bibitem[Zhang et~al.(2026{\natexlab{a}})Zhang, Wang, Gao, Su, Dai, Zhou, Lu,
  and Tang]{zhang_clap_2026}
Chubin Zhang, Jianan Wang, Zifeng Gao, Yue Su, Tianru Dai, Cai Zhou, Jiwen Lu,
  and Yansong Tang.
\newblock {CLAP}: {Contrastive} {Latent} {Action} {Pretraining} for {Learning}
  {Vision}-{Language}-{Action} {Models} from {Human} {Videos},
  2026{\natexlab{a}}.
\newblock URL \url{https://arxiv.org/abs/2601.04061}.

\bibitem[Zhang et~al.(2025{\natexlab{a}})Zhang, Pearce, Zhang, Wang, Chen,
  Shen, Zhao, and Bian]{NEURIPS2025_d7b3cef7}
Chuheng Zhang, Tim Pearce, Pushi Zhang, Kaixin Wang, Xiaoyu Chen, Wei Shen,
  Li~Zhao, and Jiang Bian.
\newblock What do latent action models actually learn?
\newblock In \emph{Advances in Neural Information Processing Systems},
  volume~38, 2025{\natexlab{a}}.
\newblock \doi{10.52202/085713-4906}.
\newblock URL
  \url{https://proceedings.neurips.cc/paper_files/paper/2025/hash/d7b3cef7c31b94a4a533db83d01a8882-Abstract-Conference.html}.

\bibitem[Zhang et~al.(2026{\natexlab{b}})Zhang, Li, Zhang, Yang, Luo, Li, Wang,
  Wang, Shao, Xu, Zhou, Shen, Jin, Xu, Ma, Liao, Lu, Shi, Wen, Zhao, Tang,
  Wang, Li, Zhu, Cheng, Xue, Zhu, Shen, and Xu]{zhang_native_2026}
Qihang Zhang, Lin Li, Luyao Zhang, Shuai Yang, Yiming Luo, Shuaiting Li, Ruilin
  Wang, Junke Wang, Jiahao Shao, Gangwei Xu, Jiaming Zhou, Yishu Shen, Yudong
  Jin, Fangyi Xu, Shuailei Ma, Jiaqi Liao, Guanxing Lu, Zifan Shi, Yongkun Wen,
  Yujie Zhao, Weixuan Tang, Xinyang Wang, Chaojian Li, Jiapeng Zhu, Ka~Leong
  Cheng, Nan Xue, Xing Zhu, Yujun Shen, and Yinghao Xu.
\newblock Native {Video}-{Action} {Pretraining} for {Generalizable} {Robot}
  {Control}, July 2026{\natexlab{b}}.
\newblock URL \url{http://arxiv.org/abs/2607.08639}.
\newblock arXiv:2607.08639 [cs.RO].

\bibitem[Zhang et~al.(2025{\natexlab{b}})Zhang, Li, Long, Zhang, Lin, Yang,
  Xie, Yang, Liu, Lin, Huang, and Zhou]{qwen3embedding}
Yanzhao Zhang, Mingxin Li, Dingkun Long, Xin Zhang, Huan Lin, Baosong Yang,
  Pengjun Xie, An~Yang, Dayiheng Liu, Junyang Lin, Fei Huang, and Jingren Zhou.
\newblock {Qwen3 Embedding}: Advancing text embedding and reranking through
  foundation models.
\newblock \emph{arXiv preprint arXiv:2506.05176}, 2025{\natexlab{b}}.

\bibitem[Zhao et~al.(2025)Zhao, Lu, Kim, Fu, Zhang, Wu, Li, Ma, Han, Finn,
  Handa, Liu, Xiang, Wetzstein, and Lin]{zhao_cot-vla_2025}
Qingqing Zhao, Yao Lu, Moo~Jin Kim, Zipeng Fu, Zhuoyang Zhang, Yecheng Wu,
  Zhaoshuo Li, Qianli Ma, Song Han, Chelsea Finn, Ankur Handa, Ming-Yu Liu,
  Donglai Xiang, Gordon Wetzstein, and Tsung-Yi Lin.
\newblock {CoT}-{VLA}: {Visual} {Chain}-of-{Thought} {Reasoning} for
  {Vision}-{Language}-{Action} {Models}, March 2025.
\newblock URL \url{http://arxiv.org/abs/2503.22020}.
\newblock arXiv:2503.22020 [cs.CV].

\bibitem[Zheng et~al.(2025)Zheng, Li, Wang, Liu, Kang, Feng, Zheng, Zou, Chen,
  Zeng, Zhang, Pang, Liu, Wang, and Zhan]{zheng2025xvla}
Jinliang Zheng, Jianxiong Li, Zhihao Wang, Dongxiu Liu, Xirui Kang, Yuchun
  Feng, Yinan Zheng, Jiayin Zou, Yilun Chen, Jia Zeng, Ya-Qin Zhang, Jiangmiao
  Pang, Jingjing Liu, Tai Wang, and Xianyuan Zhan.
\newblock {X-VLA: Soft-Prompted Transformer as Scalable Cross-Embodiment
  Vision-Language-Action Model}, 2025.
\newblock URL \url{https://arxiv.org/abs/2510.10274}.

\bibitem[Zheng et~al.(2026)Zheng, Niu, Xie, Wang, Xu, Jiang, Casta{\~{n}}eda,
  Hu, Tan, Fu, Darrell, Huang, Zhu, Xu, and Fan]{zheng2026egoscale}
Ruijie Zheng, Dantong Niu, Yuqi Xie, Jing Wang, Mengda Xu, Yunfan Jiang,
  Fernando Casta{\~{n}}eda, Fengyuan Hu, You~Liang Tan, Letian Fu, Trevor
  Darrell, Furong Huang, Yuke Zhu, Danfei Xu, and Linxi Fan.
\newblock {EgoScale}: Scaling dexterous manipulation with diverse egocentric
  human data.
\newblock \emph{CoRR}, abs/2602.16710, 2026.
\newblock \doi{10.48550/ARXIV.2602.16710}.
\newblock URL \url{https://doi.org/10.48550/arXiv.2602.16710}.

\bibitem[Zhu et~al.(2026)Zhu, Yuan, Liu, and Chen]{zhu_harp-vla_2026}
Xiang Zhu, Puzhen Yuan, Yichen Liu, and Jianyu Chen.
\newblock {HARP}-{VLA}: {Human}-{Robot} {Aligned} {Representation} {Learning}
  for {Vision}-{Language}-{Action} {Model}, May 2026.
\newblock URL \url{http://arxiv.org/abs/2605.31234}.
\newblock arXiv:2605.31234 [cs.RO].

\end{thebibliography}
